\documentclass[sn-mathphys,Numbered,iicol]{sn-jnl}

\usepackage{graphicx}%
\usepackage{multirow}%
\usepackage{amsmath,amssymb,amsfonts}%
\usepackage{amsthm}%
\usepackage{mathrsfs}%
\usepackage[title]{appendix}%
\usepackage{xcolor}%
\usepackage{textcomp}%
\usepackage{manyfoot}%
\usepackage{booktabs}%
\usepackage{algorithm}%
\usepackage{algorithmicx}%
\usepackage{algpseudocode}%
\usepackage{listings}%
\usepackage{soul}

\usepackage{lmodern}
\usepackage{anyfontsize}

\usepackage{dsfont}
\usepackage{tabularx}
\usepackage{subcaption}
\usepackage[normalem]{ulem}

\renewcommand{\epsilon}{\varepsilon}
\newcommand\ho{$\mathcal{H}_0$}
\newcommand\h{{\mathcal{H}_0}}
\newcolumntype{Y}{>{\centering\arraybackslash}X}

\begin{document}

\title[U-Flow: A U-shaped Normalizing Flow for Anomaly Detection with Unsupervised Threshold]{U-Flow: A U-shaped Normalizing Flow for Anomaly Detection with Unsupervised Threshold}



\author*[1,2]{\fnm{Matías} \sur{Tailanian}}\email{mtailanian@digitalsense.ai}

\author[1,3]{\fnm{Álvaro} \sur{Pardo}}\email{apardo@digitalsense.ai}

\author[2]{\fnm{Pablo} \sur{Musé}}\email{pmuse@fing.edu.uy}




\affil*[1]{\orgname{Digital Sense}, \orgaddress{\street{Brenda 5751}, \city{Montevideo}, \postcode{11400}, \country{Uruguay}}}
\affil[2]{\orgdiv{Facultad de Ingeniería}, \orgname{Universidad de la República}, \orgaddress{\street{Av. Julio Herrera y Reissig 565}, \city{Montevideo}, \postcode{11300}, \country{Uruguay}}}
\affil[3]{\orgdiv{Departamento de Ingeniería}, \orgname{Universidad Católica del Uruguay}, \orgaddress{\street{Av. 8 de Octubre 2738}, \city{Montevideo}, \postcode{11600}, \country{Uruguay}}}


\abstract{In this work we propose a one-class self-supervised method for anomaly segmentation in images that benefits both from a modern machine learning approach and a more classic statistical detection theory. The method consists of four phases. First, features are extracted using a multi-scale image Transformer architecture. Then, these features are fed into a U-shaped Normalizing Flow (NF) that lays the theoretical foundations for the subsequent phases. 
The third phase computes a pixel-level anomaly map from the NF embedding, and the last phase performs a segmentation based on the \textit{a contrario} framework.
This multiple hypothesis testing strategy permits the derivation of robust unsupervised detection thresholds, which are crucial in real-world applications where an operational point is needed. The segmentation results are evaluated using the Mean Intersection over Union (\textit{mIoU}) metric, and for assessing the generated anomaly maps we report the area under the Receiver Operating Characteristic curve (\textit{AUROC}), as well as the Area Under the Per-Region-Overlap curve (\textit{AUPRO}). Extensive experimentation in various datasets shows that the proposed approach produces state-of-the-art results for all metrics and all datasets, ranking first in most MVTec-AD categories, with a mean pixel-level \textit{AUROC} of 98.74\%.

Code and trained models are available at  \url{https://github.com/mtailanian/uflow}.
}




\maketitle

\section{Introduction}
\label{sec:intro}

Detecting image anomalies is a long-standing problem that has been studied for decades. In recent years, this problem has received a growing interest from the computer vision community, motivated by applications in various fields, ranging from surveillance and security to even health care. One of the most common applications, and the one that we are especially interested in, is automatizing product quality control in an industrial environment. In this case, it is often very hard (sometimes even unfeasible) to collect and label a good amount of data representing all kinds of anomalies since these are rare structures by definition. 
For this reason, the major effort on anomaly detection methods has been focused on learning only from the normal samples (i.e., anomaly-free images) and detecting the anomalies as everything that deviates from what has been learned. This approach is found in the literature under different names, usually called one-class classification, self-supervised anomaly, or novelty detection.

There is a wide literature on anomaly detection, and many approaches have been proposed. One of the most common classic approaches consists of embedding the training images in some latent space and then evaluating how far a test image is from the manifold of normal images. 
Other approaches focus on learning the probability density function of the training set. Among these, three types of generative models have become popular in anomaly detection in the last five years. On the one hand, Generative Adversarial Networks~\cite{goodfellow2014generative,anogan} implicitly learn this probability, being able only to sample from it. On the other hand, Variational Auto-Encoders~\cite{kingma2013auto,yang2020dfr} can explicitly estimate this probability density, but they can only learn an approximation, as they maximize the evidence lower bound. And finally, Normalizing Flows~\cite{dinh2014nice,dinh2016density-nvp,kingma2018glow} are able to learn the exact probability density function explicitly. Being able to do so has several advantages, as it permits to estimate the likelihood and score of how probable it is that the tested data belongs to the same training distribution. A significant advantage we exploit in this work is the possibility of developing formal statistical tests to derive unsupervised detection thresholds. This is a crucial feature for most real-world problems where a segmentation of the anomaly is needed. To this end, we propose a multiple hypothesis testing strategy based on the \textit{a contrario} framework~\cite{desolneux2007gestalt}. 

Common approaches based on VAEs~\cite{zhou2017anomaly,bergmann2018improving,gong2019memorizing} and GANs~\cite{anogan,fast-anogan-schlegl2019f,akcay2019ganomaly} have demonstrated poor performance in the industrial defect detection case, where anomalies are actually very similar to the normal parts of the images.
To enhance the performance in these cases, instead of grounding the detection on the RGB image itself, in recent years, many works have introduced a feature extraction stage using a pre-trained network, usually on ImageNet~\cite{imagenet}, and perform the anomaly detection in the feature space~\cite{kim2021semi,zheng2022focus,lee2022cfa,dfr,yu2021fastflow,csflow-rudolph2022fully}.

\begin{figure}[t]
  \centering   \includegraphics[width=1\linewidth]
  {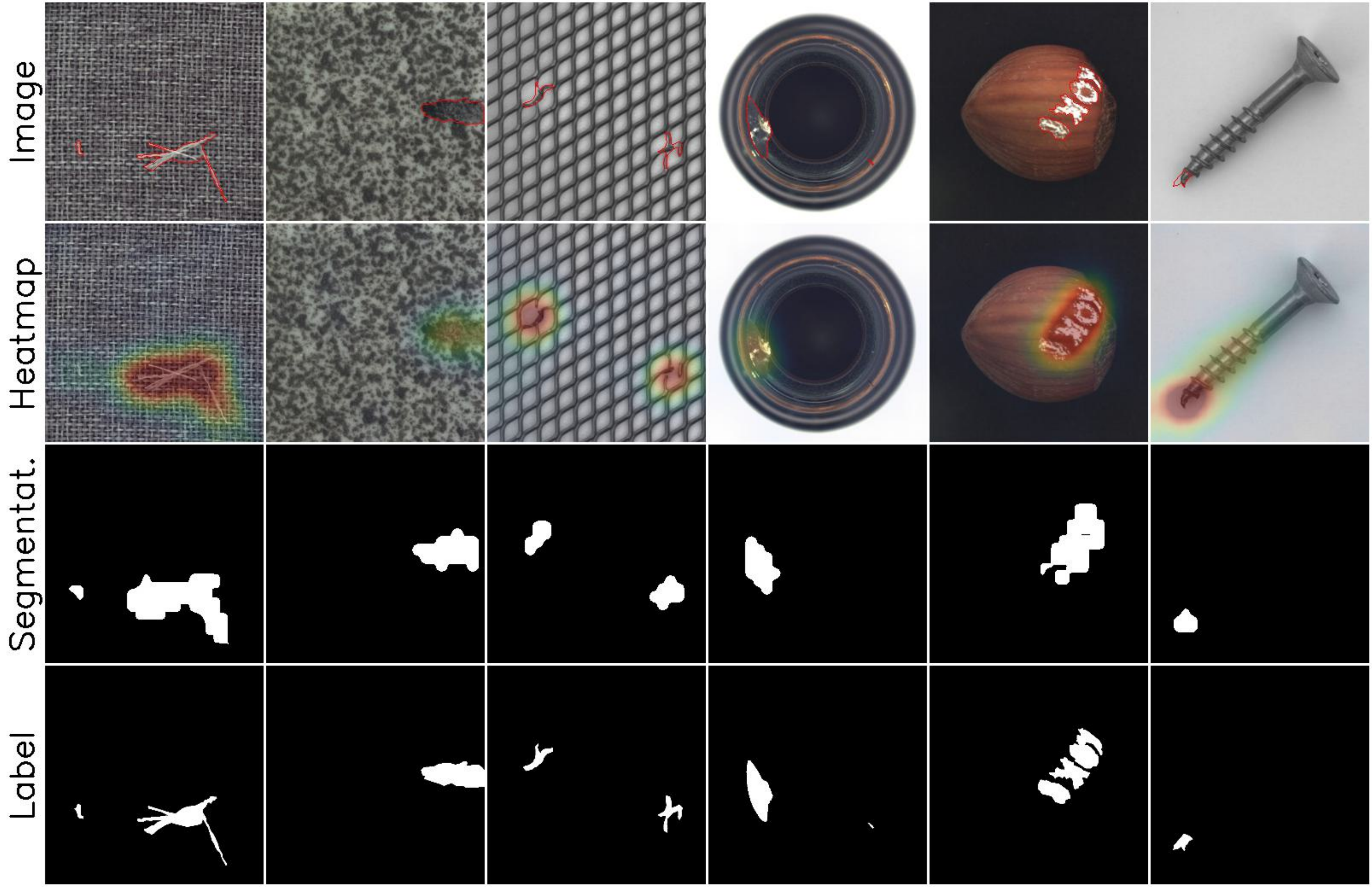}
   \caption{
   Anomalies detected with the proposed approach on MVTec-AD examples from different categories. Top row: original images with ground truth segmentation. Second row: corresponding anomaly maps. Third row: automatic segmentations. Last row: ground truth masks.
   }
   \label{fig:teaser}
\end{figure}

In this work, we propose \textbf{U-Flow}, a new self-supervised method that ensembles the features extracted by multi-scale image transformers into a U-shaped Normalizing Flow's architecture and applies the \textit{a contrario} methodology for finding an automatic segmentation of anomalies in images. Our work achieves state-of-the-art results, widely outperforming previous works in terms of \textit{mIoU} for the segmentation, and exhibiting top performance in localization metrics in terms of \textit{AUROC} and \textit{AUPRO}. The main contributions of this paper are:

\begin{itemize}
    \item For the feature extraction, we propose a multi-scale vision Transformer architecture based on the CaiT Transformers~\cite{touvron2021going-cait} pre-trained independently for each scale, which we refer to as \textbf{MS-CaiT}. Results clearly show that it better captures the multi-scale information and increases performance, comparing it both against other neural networks' feature extractors and other multi-scale Transformers. \\
    \item We leverage the well-known advantages of U-like architectures in the setting of Normalizing Flows, proposing a novel architecture and solving non-trivial technical details for merging the scales in an invertible way that better aggregates the multi-scale information elegantly and efficiently while imposing statistical mutual independence within and across scales. \\
    \item We propose, for the first time, to take advantage of the statistical properties of Normalizing Flows to derive unsupervised thresholds for anomaly detection and segmentation. More precisely, the resulting multi-scale characterization of the anomaly-free images as a white Gaussian process in the latent space allows us to propose a multiple hypothesis testing strategy that leads to an automatic detection threshold to segment anomalies, outperforming state-of-the-art methods. \\
    \item Although the method is designed for industrial defect detection tasks, we show that it also exhibits excellent performance on other benchmark datasets of different natures. \\
    \item Results are easily reproducible. Code is available on GitHub, and also integrated into \textit{Anomalib} library~\cite{anomalib}.
\end{itemize}

In short, we propose an easy-to-train method that does not require any parameter tuning or modification to the architecture (as opposed to other state-of-the-art methods), that is fast, accurate, and provides an unsupervised anomaly segmentation. Example results are shown in Figure~\ref{fig:teaser}.

The remainder of this paper is organized as follows. In Section~\ref{sec:related}, we discuss previous work related to the proposed approach. Details of the method are presented in Section~\ref{sec:method}. In Section~\ref{sec:experiments}, we compare the performance of our anomaly detection approach with a large number of state-of-the-art methods on several benchmark datasets of different nature: MVTec-AD~\cite{bergmann2019mvtec}, BeanTech~\cite{mishra2021vt-beantech}, LGG MRI~\cite{buda2019association-lggmri}, and ShanghaiTech Campus~\cite{liu2018ano_pred-stc}. An ablation study that analyzes the role of each component of the proposed architecture is presented in Section~\ref{sec:ablation}. We conclude in Section~\ref{sec:conclusion}.

\section{Related work}
\label{sec:related}

The literature on anomaly detection is vast, with numerous proposed approaches. 
In this work, we focus on one-class self-supervised learning for industrial inspection, where a key element is to learn only from anomaly-free images. Within this group, methods can be further divided into several categories in many different ways. 

A large number of methods can be classified as representation-based. These methods proceed by embedding the samples into a latent space endowed with some metric. For example, in CFA~\cite{lee2022cfa}, the authors propose a way to build features using a patch descriptor and building a scalable memory bank. A special loss function is designed to map features to a hyper-sphere and densely clusterize the representation of normal images. Also, the method can add a few abnormal samples during training that are used to enlarge the distance to the normal samples in the latent space. Similarly, in PatchSVDD~\cite{yi2020patch-svdd}, the authors construct a hierarchical encoding of image patches and encourage the network to map them to the center of a hyper-sphere in the feature space. This work later inspired~\cite{tsai2022multi-mspba}, where the authors aim to learn more representative features from normal images, using three encoders with three classifiers. In SPADE~\cite{cohen2020sub-spade}, the anomaly score is obtained by measuring the distance of a test sample to the nearest neighbors (NN) that were computed at the training stage over the anomaly-free images. Both the distance and the NN are calculated in the latent space by constructing a pyramid of features using the most common pre-trained extractors.
Additionally, PaDiM~\cite{defard2021padim} proposes to model normality by fitting a multivariate Gaussian distribution after embedding image patches with a pre-trained convolutional neural network (CNN) and base the detection on the Mahalanobis distance to normality. Then, PatchCore~\cite{roth2022towards-patchcore} further improves SPADE and PaDiM by estimating anomalies with a larger nominal context and making it less reliant on image alignment.

By using an encoder-decoder architecture such as in~\cite{yang2020dfr}, or a teacher-student scheme like~\cite{yamada2022reconstructed-rstpm}, and training the network with only anomaly-free images, the network is supposed to reconstruct normal images accurately but fail to reconstruct anomalies, as it has never seen them during training. This sweet spot when only normal structures can be generated is usually difficult to obtain, and methods tend to generate either bad-quality reconstructions or show a too-good generalization power, causing the anomalies to be well reconstructed too. The basic approach using the reconstruction error as an anomaly score is presented in~\cite{zhou2017anomaly}. Following the same idea, the authors in~\cite{bergmann2018improving} improve the detection by measuring the reconstruction error with the Structural Similarity index (SSIM), and in~\cite{gong2019memorizing} the authors limit the reconstruction power by utilizing memory modules in the latent space.

Learning from one-class data leaves some uncertainty on how the method will behave when facing anomalous data. To overcome this potential issue, some works manage to cast the problem into a self-supervised one by creating proxy tasks. For example, in~\cite{li2021cutpaste}, the authors obtain a representation of the images by randomly cutting and pasting regions over anomaly-free images to generate anomalies synthetically.

Regarding the use of Normalizing Flows (NF) in anomaly detection, a few methods have recently been proposed with impressive results. DifferNet~\cite{rudolph2021same-differnet} applies an average pooling to the features extracted with a pre-trained network and processes the resulting vectors in a one-dimensional NF, losing important contextual and spatial information. The authors try to alleviate this problem by passing different rotated image versions through the network at the cost of increasing the computational complexity. The method performs only anomaly detection, but a rough localization can also be obtained by back-propagating the negative log-likelihood through the network. CFlow~\cite{gudovskiy2022cflow} also uses a one-dimensional NF but includes the spatial information using a positional encoding as a conditional vector. On the other hand, FastFlow~\cite{yu2021fastflow} directly uses a two-dimensional NF for each scale independently and averages the results at the end. CS-Flow~\cite{csflow-rudolph2022fully} uses the same NF block as in~\cite{rudolph2021same-differnet} and~\cite{yu2021fastflow}, but for the affine layer, the information of the different scales are merged, by summing an upsampled/downsampled version of the feature volumes from the other scales. This method extends DifferNet and is also meant only for anomaly detection. However, as all the network's layers are convolutional, the spatial information is preserved, and an anomaly map can be obtained with the forward pass. The results for anomaly detection are excellent, but as the extracted features are of a small spatial size, the performance in localization could be better.

In this work, we further improve the performance of these methods for anomaly localization by proposing a multi-scale Transformer-based feature extractor and by equipping the NFs with a fully invertible U-shaped architecture that effectively integrates the information at different scales in a very natural and elegant manner. These two ingredients allow us to embed these multi-scale features in a latent space where intra and inter-scale features are guaranteed to be independent and identically distributed Gaussian random variables. Finally, we design a detection methodology from these statistical properties that leads to statistically meaningful anomaly scores and unsupervised detection thresholds. 

\section{Method}
\label{sec:method}

\begin{figure*}
  \centering
  \begin{subfigure}{1\linewidth}
    \includegraphics[width=1\linewidth]{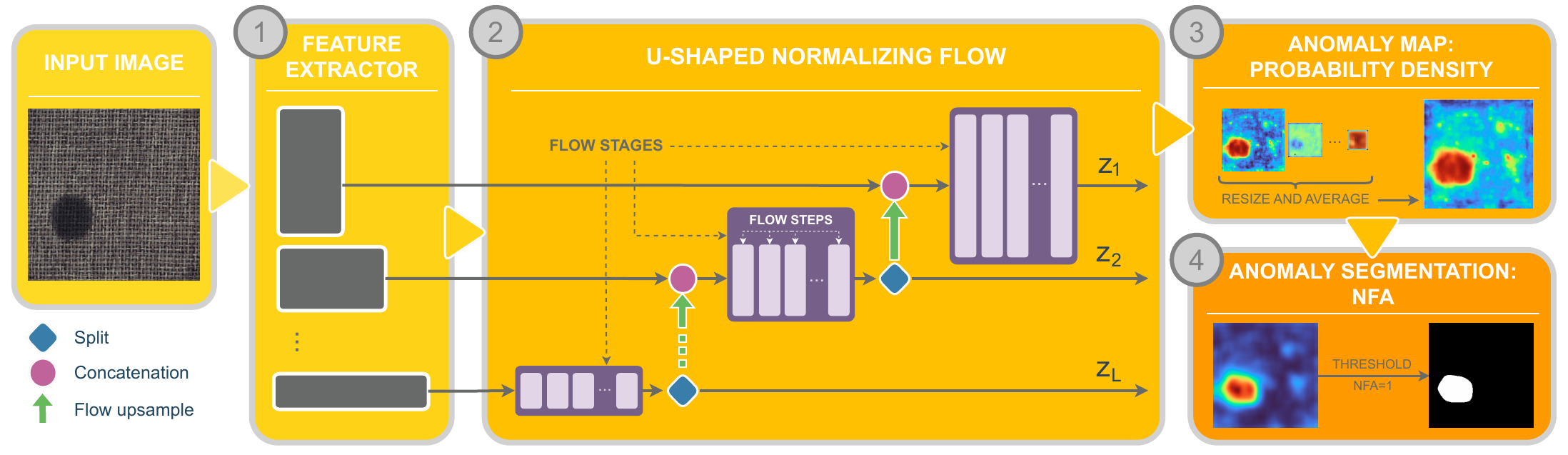}
  \end{subfigure}
  \caption{The method consists of four phases. (1) {\em Multi-scale feature extraction:} a rich multi-scale representation is obtained with MS-CaiT by combining pre-trained image Transformers acting at different image scales. (2) {\em U-shaped Normalizing Flow:} by adapting the widely used U-like architecture to NFs, a fully invertible architecture is designed. This architecture is capable of merging the information from different scales while ensuring independence in both intra- and inter-scales. To make it fully invertible, split and invertible up-sampling operations are used. 
  (3) {\em Anomaly score map generation:} an anomaly map is generated by associating a likelihood-based anomaly score to each pixel in the test image. (4) {\em Anomaly segmentation:} besides generating the anomaly map, we also propose to adapt the \textit{a contrario} framework to obtain an automatic threshold by controlling the allowed number of false alarms.}
  \label{fig:method}
\end{figure*}

The proposed method is depicted in Figure~\ref{fig:method} and consists of four main phases: 1) Feature Extraction, 2) U-shaped Normalizing Flow, 3) Anomaly score map, and 4) \textit{A contrario} anomaly segmentation. All phases are presented below.

\subsection{Phase 1: Feature Extraction}
\label{subsec:fe}

Since anomalies can emerge in various sizes and forms, collecting image information at multiple scales is essential. To do so, the standard deep learning strategy is to use pre-trained CNNs, often a VGG~\cite{vgg} or any variant of the ResNet~\cite{resnet} architectures, to extract a rich multi-scale image feature representation, by keeping the intermediate activation maps at different depths of the network. More recently, with the development of image vision Transformers, some architectures such as ViT~\cite{vit} and CaIT~\cite{touvron2021going-cait} are also being used, but in these cases, a single feature map is retrieved. The features generated by vision Transformers are proven to better compress all multi-scale information into a single feature map volume, compared to the standard CNNs. However, this notion of a unique multi-scale feature map was challenged by MViT~\cite{fan2021multiscale-mvit}, which proved that combining the standard CNNs' idea of having multi-scale feature hierarchies with the Transformer models,  the Transformers' features could be further enhanced. This work was followed by MViT2~\cite{li2021improved-mvit}, which was shown to outperform its predecessor and all other vision Transformers in various tasks. In our work, we also propose to construct a multi-scale Transformer architecture but following a different strategy. Specifically, we propose MS-CaiT, an architecture that employs CaIT Transformers at different scales, independently pre-trained on ImageNet~\cite{imagenet}, and combines them as the encoder of a U-Net~\cite{unet} architecture. Despite its simplicity, an ablation study presented in Section~\ref{sec:ablation_fe} shows that this combination strategy leads to better results than using each of the Transformers alone, far better than other ResNet variants, and even better than MViT2. We conjecture that the fact that the CaIT networks were trained independently could be beneficial because it might give the network more flexibility to concentrate on different structures in each Transformer.


As a final comment regarding the feature extraction stage, it is important to mention that, notwithstanding all the results reported in Section~\ref{sec:experiments} were obtained using exactly the same architecture configuration, the proposed architecture, and the code are agnostic to the feature extractor, and can be used with different extractors and different numbers of scales. For specific adaptation to certain situations and datasets, it could be better to use another extractor, for example, if computational power is very tight (e.g., in an industrial environment), probably at the expense of some decrease in performance.

\subsection{Phase 2: Normalizing Flow}
\label{subsec:nf}

Normalizing Flows~\cite{rezende2015variational} are a family of generative models that are trained by directly maximizing the log-likelihood of the input data and which have the particularity of learning a bijective mapping between the input distribution and the latent space. Using a series of invertible transformations, the NF can be run in both directions. The forward process embeds data into the latent space and can serve as a measure of the likelihood. The reverse process starts from a predefined distribution (usually a standard Normal distribution) and generates samples following the learned data distribution.\\

The rationale for using NFs in an anomaly detection setting is straightforward. The network is trained using only anomaly-free images, and in this way, it learns to transform the distribution of normal images into a white Gaussian noise process. At test time, when the network is fed with an anomalous image, it is expected that it will not generate a sample with a high likelihood, according to the white Gaussian noise model. Hence, a low likelihood indicates the presence of an anomaly.

This second phase is the only one that is trained. It takes the multi-scale representation generated by the feature extractor as input and performs a sequence of invertible transformations on the data using the NF framework.

State-of-the-art methods following this approach are centered on designing or trying out different multi-scale feature extractors. In this work, we further improve the approach by proposing a new feature extractor and a multi-level deep feature integration method that aggregates the information from different scales using the well-known and widely used UNet-like~\cite{unet} architecture. The U-shape comprises the feature extractor as the encoder and the NF as the decoder. We show in the ablation study of Section~\ref{sec:ablation_u} that the U-shape is a beneficial aggregation strategy that further improves the results.

The 2D NF uses only invertible operations and treats each pixel location independently, only merging the information between different channels for each pixel location. As it is implemented as one unique computational graph, we obtain a fully invertible network. Moreover, optimizing the whole flow all at once has a crucial implication: the Normalizing Flow generates a mutually independent embedding not only within each scale but also across different scales, unlike other state-of-the-art flow-based methods. 
It is worth mentioning that in most works reported so far in the literature, the anomaly score is computed first at each scale independently, using a likelihood-based estimation, and finally merged by averaging or summing the result for each scale. Because of the lack of independence between scales, these final operations, although achieving very good performance, lack a formal probabilistic interpretation. The NF architecture proposed in this work overcomes this limitation; it produces statistically independent multi-scale features, for which the joint likelihood estimation becomes trivial.\\

\noindent \textbf{Architecture.} The U-shaped NF is compounded by a number of \textit{flow stages}, each one corresponding to a different scale, whose size matches the extracted feature maps (see Figure~\ref{fig:method}). For each scale starting from the bottom, i.e., the coarsest scale, the input is fed into its corresponding \textit{flow stage}, which is essentially a sequential concatenation of \textit{flow steps}. The output of this \textit{flow stage} is then split in such a way that half of the channels are sent directly to the output of the whole graph, and the other half is up-sampled to be concatenated with the input of the next scale, and proceed in the same fashion. The up-sampling is also performed in an invertible way, as proposed in~\cite{jacobsen2018revnet}, where pixels in groups of four channels are reordered in such a way that the output volume has four times fewer channels and double width and height. 

Each \textit{flow step} has a size according to its scale and is compounded by Glow blocks~\cite{kingma2018glow}. Each step combines an Affine Coupling Layer~\cite{dinh2014nice}, a permutation using $1 \times 1$ convolutions, and a global affine transformation (ActNorm)~\cite{dinh2016density-nvp}. The Affine Coupling Layer uses a small network with the following layers: a convolution, a ReLU activation, and one more convolution. Convolutions have the same amount of filters as the input channels, and kernels alternate between $1 \times 1$ and $3 \times 3$.

To sum up, the U-shaped NF produces $L$ white Gaussian embeddings $z^1, \dots, z^L$, one for each scale $l$, $z^l \in \mathbb{R}^{C_l \times H_l \times W_l}$. Here, we denote the spatial location by $(i,j)$, and by $k$ the channel index in the latent tensors. Its elements 
\begin{equation}
    z_{ijk}^l~\sim~\mathcal{N}(0,1), 
\end{equation}
$1 \leq i \leq H_l$, $1 \leq j \leq W_l$, $1 \leq k \leq C_l$, are mutually independent for any position, channel and scale $i,j,k,l$.

\subsection{Phase 3: Likelihood-based Anomaly Score}
\label{subsec:phase3-anomaly-score}

This phase of the method is to be used at test time when computing the anomaly map. It takes as input the embeddings $\left \{z_{ijk}^l \right \}$ and produces an anomaly map based on the likelihood computation. Thanks to the statistical independence of the features produced by the U-shaped NF, the joint likelihood of a test image under the anomaly-free image model is the product of the standard normal marginals. Therefore, to build this map, we associate to each pixel $(i,j)$ in the test image, a likelihood-based Anomaly Score similar to those in~\cite{yu2021fastflow,gudovskiy2022cflow}:
\begin{equation}
    AS(i,j) = 1 - \frac{1}{L} \sum_{l=1}^L \exp \left( - \frac{1}{2C_l} \sum_{k=1}^{C_l} (\tilde{z}_{ijk}^l)^2 \right),
    \label{ec:as-likelihood}
\end{equation}
where $\tilde{z}_{ijk}^l$ is a bilinearly upsampled version of ${z}_{ijk}^l$ at the original $H\times W$ resolution. 






Note that the exponential term of the above expression does not correspond exactly to the anomaly-free likelihood since, for each scale, an average in the channel direction is computed instead of the sum. The reason for this is mainly computational: using the sum produces extremely small values close to the floating point error. Using the sum has another undesirable effect: the impact of the different scales in the total anomaly score is not equalized, as the number of channels $C_l$ may significantly differ. Although these modifications lead to an Anomaly Score that lacks a precise probabilistic interpretation, this computation produces high-quality anomaly maps and increases performance in practice.

In addition, in the following section, we introduce a more formal and statistically meaningful result, using the \textit{a contrario} theory, that permits to derive an unsupervised detection threshold on the {\em Number of False Alarms (NFA)}, and produces an anomaly segmentation mask.

\subsection{Phase 4: \textit{A contrario} anomaly segmentation}
\label{sec:nfa}

Besides generating the anomaly map discussed in the previous section, we present a new method for computing an automatic segmentation of the anomalies based on the \textit{a contrario} approach. In this approach, an estimate of the Number of False Alarms (NFA) is associated to each anomaly candidate, and detections are obtained by thresholding the NFA~\cite{desolneux2007gestalt}. The proposed method has no parameters to tune and uses level sets organized into a tree structure to segment the anomalies automatically.

\subsubsection{Number of False Alarms}
\label{subsec:nfa}

The {\em a contrario} framework~\cite{desolneux2007gestalt} is a multiple hypothesis testing methodology that has been successfully applied to derive unsupervised statistical detection thresholds in a wide variety of detection problems, such as alignments for line segment detection~\cite{von2008straight}, clustering~\cite{cao2007unified}, image forgery detection~\cite{marina}, and even anomaly detection~\cite{grosjean2009contrario, davy2018reducing-thibaud-noise,tailanian2022contrario-book}, to name a few. In~\cite{r3_ehret2019image}, the authors revisit several anomaly detection methods from the perspective of the \textit{a contrario} framework to derive unsupervised statistical detection thresholds. The approach we propose in this work follows the same rationale, as it exploits the statistical independence of the U-Flow embeddings to derive anomaly detection thresholds.

The \textit{a contrario} framework is based on the non-accidentalness principle~\cite{lowe1985perceptual}. Given that we do not usually know what the anomalies look like, it focuses on modeling normality by defining a null hypothesis (\ho), also called {\em background model}. Relevant structures are detected as large deviations from this model by evaluating how likely it is that an observed structure or event $E$ would happen under \ho. 
One of its most useful characteristics is that it provides an easily interpretable detection threshold based on controlling the NFA, defined as
\begin{equation}
    \text{NFA}(E) = N_T \mathrm{Pr_\h}(E).
    \label{ec:nfa}
\end{equation}
Here $N_T$ denotes the number of events that are tested, and $\mathrm{Pr_\h}(E)$ 
is the probability of the event $E$ being a realization of the background model. Besides providing a robust detection threshold, as it will be shown, the NFA value itself has a clear statistical meaning: it is an estimate of the expected number of times, among the tests that are performed, that such tested event could be generated by the background model~\cite{desolneux2007gestalt}. 
In other words, an event $E$ is $\epsilon$-meaningful if its expected number of occurrences is less than $\epsilon$ under the normality assumption. Consequently, $\epsilon$ provides an upper bound on the expected number of false detections we can obtain on anomaly-free images. A low NFA value means that the observed pattern is too unlikely to be generated by the background model and, therefore, indicates a meaningful anomaly. \\ 

As the events $E$ corresponding to anomalies can have arbitrary sizes and shapes, we must test all possible configurations of connected sets of pixels within every image. But this is computationally very inefficient and possibly even impossible in a reasonable amount of time. Therefore, we propose to use the anomaly map to select the regions to be tested from the set of connected components of its level sets. These components are naturally ordered by inclusion, providing a convenient hierarchical representation that can be organized in a tree structure.

\subsubsection{Detecting anomalies from the level sets of the anomaly map}
\label{sec:tree}

\begin{figure}[t]
  \begin{subfigure}[t]{0.47\columnwidth}
  \centering
    \includegraphics[width=.85\linewidth]{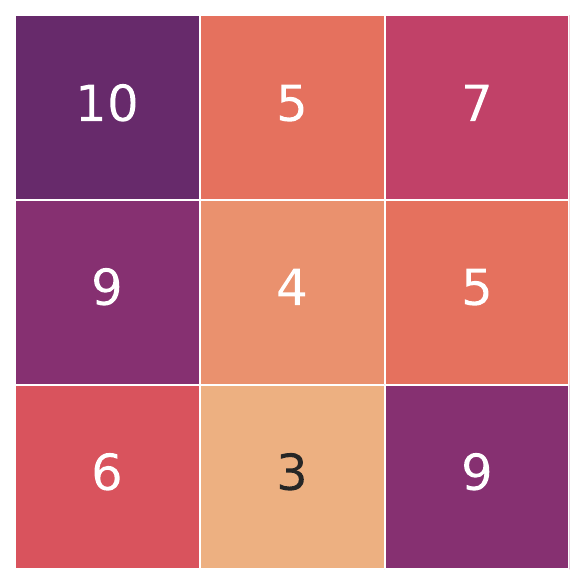}
    \caption{Example image $u$ used to build the tree. Each pixel has its value annotated.}
    \label{fig:tree_nfa_a}
  \end{subfigure}
  \hfill 
  \begin{subfigure}[t]{0.47\columnwidth}
  \centering
    \includegraphics[width=.85\linewidth]{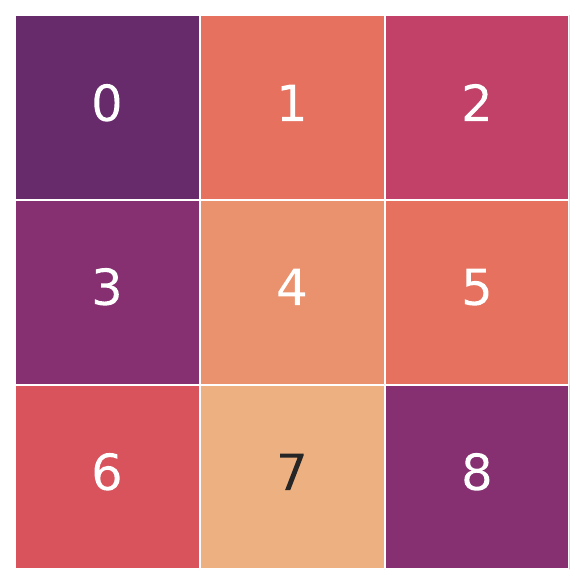}
    \caption{Same image $u$, annotated with the raveled index of each pixel.}
    \label{fig:tree_nfa_b}
  \end{subfigure}
  \\ 
  \begin{subfigure}[t]{1\columnwidth}
  \centering
    \vspace{-10pt}
    \includegraphics[width=\linewidth]{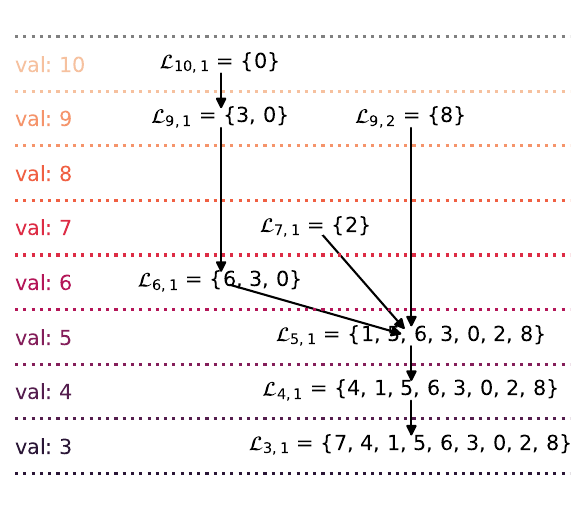}
    \caption{Example tree of connected components built from the upper level sets of (a). Each node is represented by a set of indexes from (b), and each row corresponds to a different level set, according to the values of (a).}
    \label{fig:tree_nfa_c}
  \end{subfigure}
\caption{Tree of connected components of the upper level sets: the hierarchical representation based on the level sets used to retrieve the most significant regions to be tested for anomaly segmentation.}
\label{fig:tree_nfa}
\end{figure}

In image processing, level sets are the basis for a wide variety of morphological filters, and, as they are closely related to object boundaries, they have been used for edge detection~\cite{desolneux2001edge, cao2005extracting}, registration~\cite{ls-monasse2000fast,muse2006contrario}, image quantization~\cite{ls-ballester2003tree}, and segmentation~\cite{ls-xu2012context}, among others. All the information of a gray-valued image $u(i,j)$ is contained in a set of binary images obtained by thresholding at different levels. Since no information is lost, level sets provide a complete representation: any image can be easily reconstructed from the whole family of its (upper) level sets~\cite{serra1982image}
\begin{equation}
\mathcal{L}_\lambda = \left\lbrace (i,\,j) \; | \; u(i,\,j) \geq \lambda \right\rbrace.
\end{equation}
We denote by $\mathcal{L}_{\lambda,c}$ the $c$-th connected component of $\mathcal{L}_{\lambda}$.

Level sets naturally define a hierarchical representation, ordered by their geometrical inclusion. In the case of the upper level sets, each connected component defined by a certain level is included in another connected component defined by a lower level. Therefore, they can be naturally embedded in a tree representation.

We treat all scales produced by the U-Flow independently and use these properties to build one \textit{tree of connected components} for each scale. For simplicity, in this section, we present the method for one scale, and later in Section~\ref{sec:NFA_computation}, we explain how to combine the results from different scales into a final set of detections. In the following, we omit the superscript $l$ used to denote the embedding scale.

Specifically, for each scale, we propose to use the upper level sets of an image defined from the NF embeddings as
\begin{equation}
    u(i,j)=\sum_{k=1}^{C}(z_{ijk})^2,
    \label{ec:def_of_u}
\end{equation}
where $C$ is the number of channels of the embedding. Note that for building the tree of connected components, as we are only interested in the level sets, using $u(i,j)$ is equivalent to using the Anomaly Score~\eqref{ec:as-likelihood} for a single scale $l$ since this score is an increasing function of $u(i,j)$. 
%
%
%
%
%
%
%

Figure~\ref{fig:tree_nfa} shows a toy example that illustrates the procedure. Figure~\ref{fig:tree_nfa_a} represents the image $u$, with values $u(i,j)$ denoted by numbers inside each pixel. Figure~\ref{fig:tree_nfa_b} shows the raveled indexes that serve as labels for the tree nodes shown in Figure~\ref{fig:tree_nfa_c}. The tree is built so that the nodes are given by the connected components of the level set $\mathcal{L}_{\lambda,c}$, and the edges represent the inclusion, connecting level set components from two different levels.

The tree of connected components allows us to significantly reduce the number of image regions we effectively test. Instead of computing the NFA of all possible  $N_T$ connected regions of any shape and size, which would be intractable, we will limit this computation to the connected components of the upper level sets in the tree. By construction, these regions are expected to be good candidates for anomalies. 

\begin{figure}[t]
  \centering
  \includegraphics[width=1\linewidth]{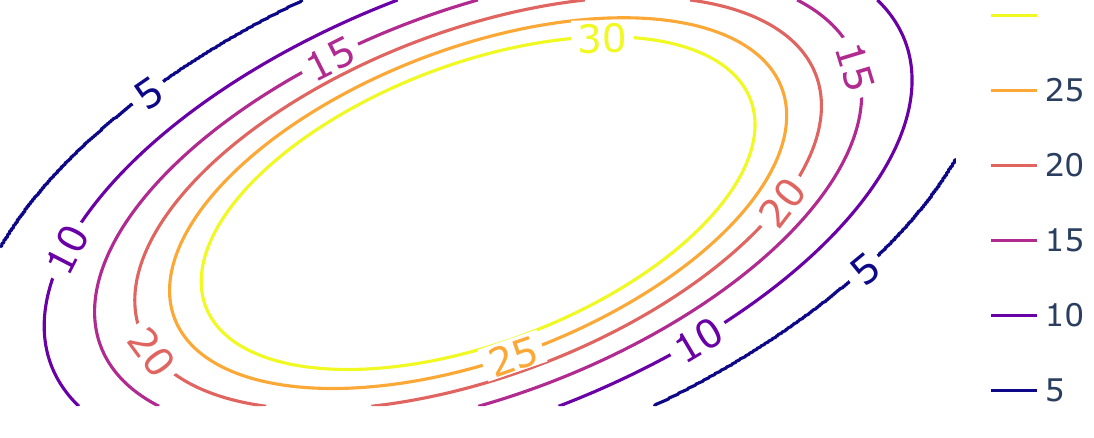}
   \caption{Example level lines of a branch of the tree of connected components.}
   \label{fig:level_lines}
\end{figure}

To compute the NFA of a connected component $\mathcal{L}_{\lambda,c}$ in the tree, we need to evaluate the Probability of False Alarm (PFA), i.e., the probability of occurrence of a region like the observed $\mathcal{L}_{\lambda,c}$ under the background model of normality. In the absence of anomaly, the embedding produced by the U-Flow model is $z_{ijk}^l \sim \mathcal{N}(0,1), \; \text{i.i.d.}$ Therefore, variables $u(i,j)$ as defined in~\eqref{ec:def_of_u} are identically distributed as a \textit{Chi-Squared} distribution of order $C$. Then, as the minimum pixel value in the connected component $\mathcal{L}_{\lambda,c}$ is $\lambda$, it follows that its probability of false alarms is given by
\begin{align}
    \text{PFA}(\mathcal{L}_{\lambda,c}) & = 
    \mathrm{Pr}\left(\min_{(i,j) \in \mathcal{L}_{\lambda,c}} u(i,j) \geq \lambda\right) \nonumber \\
    & = \left( 1 - \underset{\mathcal{X}^2(C)}{\textit{CDF}} \left( \lambda \right) \right)^{|\mathcal{L}_{\lambda,c}|},
    \label{ec:nfa_cc}
\end{align}
where $|\mathcal{L}_{\lambda,c}|$ denotes the number of pixels in $\mathcal{L}_{\lambda,c}$.
This measure of meaningfulness of a region presents a natural balance between growing the regions (and with that increasing $|\mathcal{L}_{\lambda,c}|$), and reducing the minimum value of its pixels (i.e. reducing $\lambda$). This balance is desirable to delineate the abnormal regions well.

For efficiency purposes, instead of computing the $\textit{CDF}$ in~\eqref{ec:nfa_cc}, we use the {Chernoff bound}~\cite{chernoff1952measure} for a $\mathcal{X}^2(C_l)$ distribution obtaining the final base 10 logarithm of the $\text{PFA}$ value:
\begin{equation}
\begin{split} \log(\text{PFA} (\mathcal{L}&_{\lambda,c})) = \\ 
     =|\mathcal{L}_{\lambda,c}| & \frac{C_l}{2 \ln(10)} \left( 1 + \ln \left( \frac{\lambda}{C_l} \right) - \frac{\lambda}{C_l} \right).
\end{split}
\end{equation}

Once we have computed the $\log(\text{PFA})$ values of all connected components in the tree, to obtain the final regions, we iteratively perform the following \textit{prune} and \textit{merge} procedures until convergence (until the tree does not change anymore).\\

\noindent\textbf{Procedure 1: $\text{PFA}$ Prune}. 
This procedure aims to filter a set of nested connected components, keeping only the most significant one. To do so, we directly use the $\log(\text{PFA})$ value of each node (each connected component), and select the lowest one. Starting from each leaf, we identify a connected section of the tree where all nodes have exactly one predecessor (except for the starting leaf itself). The concentric level lines defined by such a branch are illustrated by the example in Figure~\ref{fig:level_lines}. In other words, we identify which of these connected components is the most significant one, as it may better delineate the anomalous region. After determining which node to preserve, the tree is pruned so that just the chosen node is kept, and all other branch nodes are removed. In the example of Figure~\ref{fig:tree_nfa_c}, we would consider the branch $b_1 = \{\mathcal{L}_{10,1},\,\mathcal{L}_{9, 1},\,\mathcal{L}_{6, 1}\}$, evaluate the $\log(\text{PFA})$ value for each node, and prune the tree, keeping only the one with lowest value, i.e. the most significant region, denoted as $\mathcal{L}_{b_1}$.\\

\noindent\textbf{Procedure 2: $\text{PFA}$ Merge}. 
The second procedure consists of merging leaf nodes with the same successor in case the latter is more significant than all others. In this case, all leaf nodes are removed from the tree, and we only keep their successor. 
In the example of Figure~\ref{fig:tree_nfa_c}, this procedure would analyze if the node $\mathcal{L}_{5,1}$ (the successor) is more significant than $\mathcal{L}_{b_1}$ (the leaf predecessor node resulting from \textit{Procedure 1}), $\mathcal{L}_{9,2}$, and $\mathcal{L}_{7,1}$. In case $\mathcal{L}_{5,1}$ has the lowest $\log(\text{PFA})$ value between all these nodes, the other nodes are removed from the tree. Otherwise, this procedure makes no changes to the tree.\\

Note that in the case some actual merge is performed in \textit{Procedure 2}, it could generate a new branch with more than one node, starting from a new leaf, where all nodes in the branch have only one predecessor (except the starting leaf itself). Therefore, we iteratively apply both procedures until no more changes are performed over the tree. At the end of this iterative process, we use the resulting leaf nodes of the tree as the final regions and assign them the $\log(\text{PFA})$ value of the corresponding connected component, generating a $\log(\text{PFA})$ heatmap per scale.

\subsubsection{log(NFA) computation}
\label{sec:NFA_computation}
The scales are merged by upsampling the resulting $\log(\text{PFA})$ heatmaps at different scales $l$, up to the original input size $(H,\,W)$, and keeping the minimum value at each pixel position $(i,\,j)$:
\begin{equation}
    \log\left(\text{PFA}(i,j)\right) = \min_l\left\{ \overset{H,W}{\uparrow}\left(\log\left(\text{PFA}^l\right)\right)(i,j) \right\},
\end{equation}
where $\overset{H,W}{\uparrow}\left( \cdot \right)$ represents the bilinear upsample up to the original size.\\

Finally, for the number of tests $N_T$, we need to quantify all possible regions, or arrangements of connected pixels of any size and shape, located in any part of the image. Considering 4-connectivity, these groups of connected pixels correspond to the figures called \textit{polyominoes}, and a good approximation for the number of polyominoes of $r$ pixels is given by $\alpha \beta^{r} / r$, with $\alpha=0.316915$, and $\beta=4.062570$ (see~\cite{jensen2000statistics,von2021ground}). Therefore, considering all possible region sizes and shapes in all possible spatial positions of all scales, results in
\begin{equation}
    N_T =
    \sum_l H_l \, W_l \sum_{r=1}^{H_l W_l} \alpha \frac{\beta ^ r}{r}. 
\end{equation}
Finally, the $\log(\text{NFA})$ value is given by
\begin{equation}
    \log(\text{NFA}(i,\,j)) = \log(N_T) + \log(\text{PFA}(i,\,j)).
\end{equation}

\begin{table*}[t]

  \small
  
  \centering
  \begin{tabularx}{1\textwidth}{@{}cYYYYYYYYYY}
\toprule
\textbf{Category} & \begin{tabular}{@{}c@{}}\textbf{Patch}\\ \textbf{SVDD} \end{tabular} & \begin{tabular}{@{}c@{}}\textbf{SPADE}\end{tabular} & \begin{tabular}{@{}c@{}}\textbf{PaDiM} \end{tabular} & \begin{tabular}{@{}c@{}}\textbf{Cut}\\ \textbf{Paste} \end{tabular} & \begin{tabular}{@{}c@{}}\textbf{Patch}\\ \textbf{Core} \end{tabular}& \begin{tabular}{@{}c@{}}\textbf{PEFM}\end{tabular} & \begin{tabular}{@{}c@{}}\textbf{Fast}\\ \textbf{Flow} \end{tabular} & \begin{tabular}{@{}c@{}}\textbf{CFlow}\end{tabular} & \begin{tabular}{@{}c@{}}\textbf{CS-}\\ \textbf{Flow}\end{tabular} & \begin{tabular}{@{}c@{}}\textbf{U-Flow}\\ \textbf{(ours)}\end{tabular} \\
\midrule
\textbf{Carpet}      &  92.60  &  97.50  &  99.10  &  98.30           &  98.90  &  99.00  &  99.40           &  99.25           & 95.23 &  \textbf{99.42} \\
\textbf{Grid}        &  96.20  &  93.70  &  97.30  &  97.50           &  \textbf{98.70}  &  98.48  &  98.30           &  \textbf{98.99}  & 94.46 &  98.49          \\
\textbf{Leather}     &  97.40  &  97.60  &  98.90  &  99.50           &  99.30  &  99.24  &  99.50           &  \textbf{99.66}  & 89.32 &  \textbf{99.59}          \\
\textbf{Tile}        &  91.40  &  87.40  &  94.10  &  90.50           &  95.60  &  95.19  &  96.30           &  \textbf{98.01}  & 90.71 &  \textbf{97.54}          \\
\textbf{Wood}        &  90.80  &  88.50  &  94.90  &  95.50           &  95.00  &  95.27  &  \textbf{97.00}           &  96.65           & 86.29 &  \textbf{97.49} \\
\midrule
\textbf{Av. texture} &  93.68  &  92.94  &  96.86  &  96.26           &  97.50  &  97.44  &  \textbf{98.10}           &  \textbf{98.51}  & 91.20 &  \textbf{98.51} \\
\midrule
\textbf{Bottle}      &  98.10  &  98.40  &  98.30  &  97.60           &  98.60  &  98.11  &  97.70           &  \textbf{98.98}  & 88.52 &  \textbf{98.65}          \\
\textbf{Cable}       &  96.80  &  97.20  &  96.70  &  90.00           &  98.40  &  96.58  &  98.40           &  97.64           & 96.08 &  \textbf{98.61} \\
\textbf{Capsule}     &  95.80  &  99.00  &  98.50  &  97.40           &  98.80  &  97.94  &  \textbf{99.10}  &  98.98           & 98.13 &  \textbf{99.02}          \\
\textbf{Hazelnut}    &  97.50  &  \textbf{99.10}  &  98.20  &  97.30           &  98.70  &  98.78  &  \textbf{99.10}           &  98.89           & 96.13 &  \textbf{99.30} \\
\textbf{Metal nut}   &  98.00  &  98.10  &  97.20  &  93.10           &  98.40  &  96.89  &  98.50           &  \textbf{98.56}           & 95.98 &  \textbf{98.82} \\
\textbf{Pill}        &  95.10  &  96.50  &  95.70  &  95.70           &  97.10  &  96.67  &  \textbf{99.20}           &  98.95           & 95.83 &  \textbf{99.35} \\
\textbf{Screw}       &  95.70  &  98.90  &  98.50  &  96.70           &  \textbf{99.40}  &  98.93  &  \textbf{99.40}           &  98.86           & 98.28 &  \textbf{99.49} \\
\textbf{Toothbrush}  &  98.10  &  97.90  &  98.80  &  98.10           &  98.70  &  98.28  &  \textbf{98.90}           &  \textbf{98.93}  & 97.39 &  98.79          \\
\textbf{Transistor}  &  97.00  &  94.10  &  97.50  &  93.00           &  96.30  &  96.58  &  97.30           &  \textbf{97.99}  & 96.34 &  \textbf{97.87}          \\
\textbf{Zipper}      &  95.10  &  96.50  &  98.50  &  \textbf{99.30}  &  98.80  &  98.29  &  98.70           &  \textbf{99.08}           & 95.82 &  98.60          \\
\midrule
\textbf{Av. objects} &  96.72  &  97.57  &  97.79  &  95.82           &  98.32  &  97.71  &  98.63           &  \textbf{98.69}           & 95.85 &  \textbf{98.85} \\
\midrule
\midrule
\textbf{Av. total}   &  95.71  &  96.03  &  97.48  &  95.97           &  98.05  &  97.62  &  98.45           &  \textbf{98.63}           & 94.30 &  \textbf{98.74} \\
\bottomrule
  \end{tabularx}
  \caption{MVTec-AD results: pixel-level \textit{AUROC}. The two best results for each row are in bold. Comparison with the best performing methods: Patch SVDD~\cite{yi2020patch-svdd}, SPADE~\cite{cohen2020sub-spade}, PaDiM~\cite{defard2021padim}, Cut Paste~\cite{li2021cutpaste}, Patch Core ~\cite{roth2022towards-patchcore}, PEFM~\cite{wan2022position-pefm}, Fast Flow~\cite{yu2021fastflow}, CFlow~\cite{gudovskiy2022cflow}, and CS-Flow~\cite{csflow-rudolph2022fully}. Our method outperforms all previous methods, with an average value of 98.74\%.}
  \label{tab:pixel_auroc}
\end{table*}

\begin{table*}[t]

  \small

  \centering

  \begin{tabularx}{1\textwidth}{cYYYYYYc}
    \toprule
    
    \textbf{Category}    &  \textbf{SPADE} & \textbf{PaDiM} &  \textbf{PEFM}  & \textbf{FastFlow}& \textbf{CFlow} & \textbf{CS-Flow} & \begin{tabular}{@{}c@{}}\textbf{U-Flow}\\ \textbf{(ours)}\end{tabular} \\
    \midrule
    \textbf{Carpet}      &  94.70          & 96.20          &  96.75          & 97.30          & \textbf{97.70} & 83.55          & \textbf{98.49}  \\
    \textbf{Grid}        &  86.70          & 94.60          &  \textbf{97.21} & 94.79          & \textbf{96.08} & 78.83          & 95.46           \\
    \textbf{Leather}     &  97.20          & 97.80          &  98.91          & \textbf{99.16} & \textbf{99.35} & 81.05          & 98.40           \\
    \textbf{Tile}        &  75.90          & 86.00          &  91.10          & 86.90          & \textbf{94.34} & 72.97          & \textbf{93.01}  \\
    \textbf{Wood}        &  87.40          & 91.10          &  95.77          & 94.14          & \textbf{95.79} & 60.49          & \textbf{95.80}  \\
    \midrule
    \textbf{Av. Texture} & 88.38           & 93.14          & 95.95           & 94.46          & \textbf{96.65} & 75.38          & \textbf{96.23}  \\
    \midrule
    \textbf{Bottle}      &  95.50          & 94.80          &  \textbf{95.92} & 91.55          & \textbf{96.80} & 64.14          & 95.64           \\
    \textbf{Cable}       &  90.90          & 88.80          &  \textbf{97.73} & \textbf{94.38} & 93.53          & 84.07          & 93.73           \\
    \textbf{Capsule}     &  \textbf{93.70} & 93.50          &  92.11          & 91.22          & 93.40          & 89.30          & \textbf{95.02}  \\
    \textbf{Hazle Nut}   &  95.40          & 92.60          &  \textbf{97.99} & 97.31          & 96.68          & 82.77          & \textbf{97.35}  \\
    \textbf{Metal Nut}   &  \textbf{94.40} & 85.60          &  93.88          & 91.05          & 91.65          & 76.93          & \textbf{94.32}  \\
    \textbf{Pill}        &  94.60          & 92.70          &  96.18          & \textbf{96.80} & 95.39          & 75.61          & \textbf{97.18}  \\
    \textbf{Screw}       &  \textbf{96.00} & 94.40          &  95.73          & 67.00          & 95.30          & 92.21          & \textbf{97.33}  \\
    \textbf{Toothbrush}  &  93.50          & 93.10          &  \textbf{96.21} & 92.22          & \textbf{95.06} & 78.48          & 88.80           \\
    \textbf{Transistor}  &  87.40          & 84.50          &  90.84          & 91.76          & 81.40          & \textbf{91.87} & \textbf{91.88}  \\
    \textbf{Zipper}      &  92.60          & 95.90          &  \textbf{96.45} & 95.19          & \textbf{96.60} & 85.84          & 95.47           \\
    \midrule
    \textbf{Av. Objects} & 93.40           & 91.59          & \textbf{95.30}  & 91.05          & 93.58          & 82.12          & \textbf{94.67}  \\
    \midrule
    \midrule
    \textbf{Av. Total}   &  91.73          & 92.11          &  \textbf{95.52} & 92.18          & 94.60          & 79.87          & \textbf{95.19}  \\
    \bottomrule
  
  \end{tabularx}
  \caption{MVTec-AD results: pixel-level \textit{AUPRO}. The two best results for each category are shown in bold. Comparison with the best-performing method available: SPADE~\cite{cohen2020sub-spade}, PaDiM~\cite{defard2021padim}, PEFM~\cite{wan2022position-pefm}, and the best flow-based methods: FastFlow~\cite{yu2021fastflow}, CFlow~\cite{gudovskiy2022cflow}, and CS-Flow~\cite{csflow-rudolph2022fully}. Patch Core~\cite{roth2022towards-patchcore} is excluded from the table as it only reports a total average of 93.50\%. Our method also achieves state-of-the-art performance for this metric, even obtaining the best result for some categories.} 
  \label{tab:pixel_aupro}
\end{table*}

\newcommand\bb[1]{\textbf{#1}}

\begin{table*}[t]

  \small

  \centering
  \begin{tabularx}{1\textwidth}{c | YYYY | YYYY}
\toprule
& \multicolumn{4}{c|}{\textbf{Oracle threshold}} & \multicolumn{4}{c}{\textbf{Fair threshold}} \\
\cmidrule{2-9} 
\textbf{Category}& \begin{tabular}{@{}c@{}}\textbf{Fast}\\ \textbf{Flow}\end{tabular} & \begin{tabular}{@{}c@{}}\textbf{C-} \\ \textbf{Flow} \end{tabular} & \begin{tabular}{@{}c@{}}\textbf{CS-} \\ \textbf{Flow} \end{tabular} & \textbf{Ours} & \begin{tabular}{@{}c@{}}\textbf{Fast} \\ \textbf{Flow} \end{tabular} & \begin{tabular}{@{}c@{}}\textbf{C-} \\ \textbf{Flow} \end{tabular} & \begin{tabular}{@{}c@{}}\textbf{CS-} \\ \textbf{Flow} \end{tabular} & \textbf{Ours} \\
\midrule
\textbf{Carpet}      & 0.5451      & 0.4519 & 0.3521 & \bb{0.5664} & 0.4267      & 0.2393      & 0.3420 & \bb{0.5663} \\
\textbf{Grid}        & \bb{0.4664} & 0.4077 & 0.3498 & 0.4316      & 0.2235      & 0.3413      & 0.3488 & \bb{0.4307} \\
\textbf{Leather}     & \bb{0.5483} & 0.4818 & 0.2986 & 0.4242      & \bb{0.5241} & 0.3096      & 0.2823 & 0.4183      \\
\textbf{Tile}        & 0.5564      & 0.3705 & 0.4906 & \bb{0.6229} & 0.4342      & 0.3105      & 0.2544 & \bb{0.6188} \\
\textbf{Wood}        & 0.4792      & 0.3767 & 0.3400 & \bb{0.5523} & 0.4584      & 0.2433      & 0.0806 & \bb{0.5427} \\
\midrule
\textbf{Av. Texture} & 0.5191      & 0.4177 & 0.3662 & \bb{0.5195} & 0.4134      & 0.2888      & 0.2616 & \bb{0.5154} \\
\midrule
\textbf{Bottle}      & 0.5750      & 0.6503 & 0.4098 & \bb{0.6617} & 0.4053      & 0.6434      & 0.3935 & \bb{0.6564} \\
\textbf{Cable}       & 0.5375      & 0.4009 & 0.5018 & \bb{0.6514} & 0.4807      & 0.2299      & 0.3668 & \bb{0.6491} \\
\textbf{Capsule}     & 0.3439      & 0.3040 & 0.2739 & \bb{0.4182} & 0.3160      & 0.3002      & 0.2669 & \bb{0.4124} \\
\textbf{HazleNut}    & 0.4723      & 0.5610 & 0.4654 & \bb{0.6546} & 0.4671      & 0.4831      & 0.2747 & \bb{0.6511} \\
\textbf{MetalNut}    & 0.5801      & 0.3981 & 0.4631 & \bb{0.6307} & 0.5801      & 0.2998      & 0.4165 & \bb{0.6267} \\
\textbf{Pill}        & 0.4616      & 0.3121 & 0.2324 & \bb{0.4815} & 0.4051      & 0.3096      & 0.2260 & \bb{0.4803} \\
\textbf{Screw}       & 0.2562      & 0.3039 & 0.3137 & \bb{0.3811} & 0.2375      & 0.2755      & 0.3049 & \bb{0.3726} \\
\textbf{Toothbrush}  & 0.3625      & 0.3347 & 0.3160 & \bb{0.4188} & 0.3007      & 0.2416      & 0.3016 & \bb{0.4121} \\
\textbf{Transistor}  & 0.6337      & 0.6000 & 0.6622 & \bb{0.7381} & 0.0597      & 0.2981      & 0.6516 & \bb{0.7354} \\
\textbf{Zipper}      & \bb{0.5772} & 0.5001 & 0.4092 & 0.5460      & 0.4184      & 0.2403      & 0.4082 & \bb{0.5395} \\
\midrule
\textbf{Av. Object}  & 0.4800      & 0.4365 & 0.4048 & \bb{0.5582} & 0.3671      & 0.3322      & 0.3611 & \bb{0.5536} \\
\midrule
\midrule
\textbf{Total}       & 0.4930      & 0.4302 & 0.3919 & \bb{0.5453} & 0.3825      & 0.3177      & 0.3279 & \bb{0.5408} \\
\bottomrule
  \end{tabularx}
  \caption{Segmentation \textit{mIoU} comparison for MVTec-AD, with the best flow-based methods in the literature: FastFlow~\cite{yu2021fastflow}, CFlow~\cite{gudovskiy2022cflow}, and CS-Flow~\cite{csflow-rudolph2022fully}, for the oracle-like and fair thresholds defined in Section~\ref{sec:segmentation-results}. Our method largely outperforms all others, and even exhibits a better performance comparing the proposed automatic threshold with their oracle-like threshold.}
  \label{tab:miou}
\end{table*}

\begin{figure*}[t]
  \centering
   \includegraphics[width=1\linewidth]
   {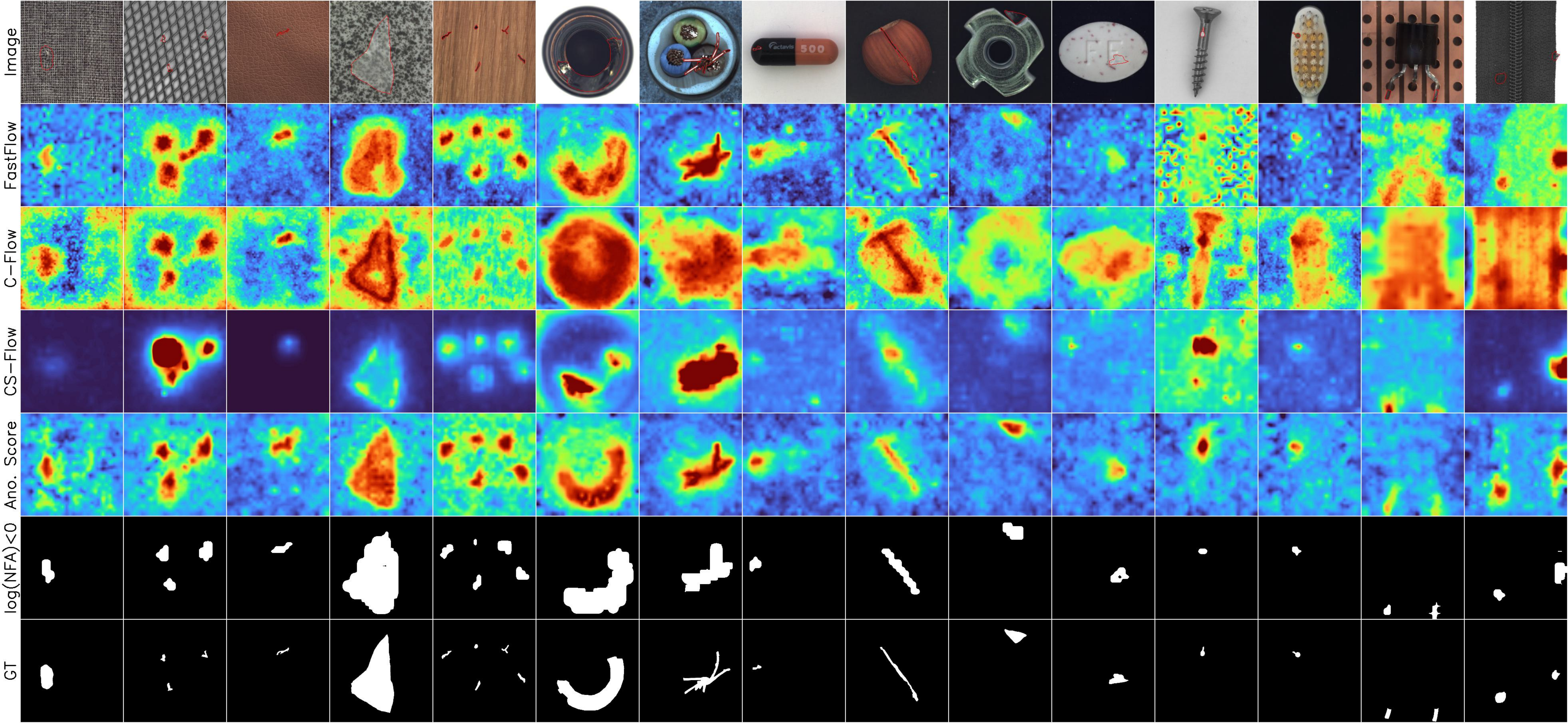}
   \caption{Example results for all MVTec categories. The first row shows the example images with the ground truth over-imposed in red. The results for FastFlow, CFlow, and CS-Flow are shown in the second, third, and fourth rows.
   The next two rows correspond to our method: the anomaly score defined in~\eqref{ec:as-likelihood}, and the segmentation obtained with the automatic threshold $\log (\text{NFA}) < 0$. The last row presents the segmentation masks for an easy comparison. While other methods achieve a very good performance, in some cases, they present artifacts and over-estimated anomaly scores. Our anomaly score achieves very good visual and numerical results, spotting anomalies with high confidence. Finally, the segmentation with the automatic threshold on the NFA is also able to spot and accurately segment the anomalies. All detections of these examples exhibit very low $\log(\text{NFA})$ values, ranging from -50 to -1515.}
   \vspace{-10pt}
   \label{fig:results}
\end{figure*}

\begin{figure*}[ht]
  \centering
  \includegraphics[width=1\linewidth]{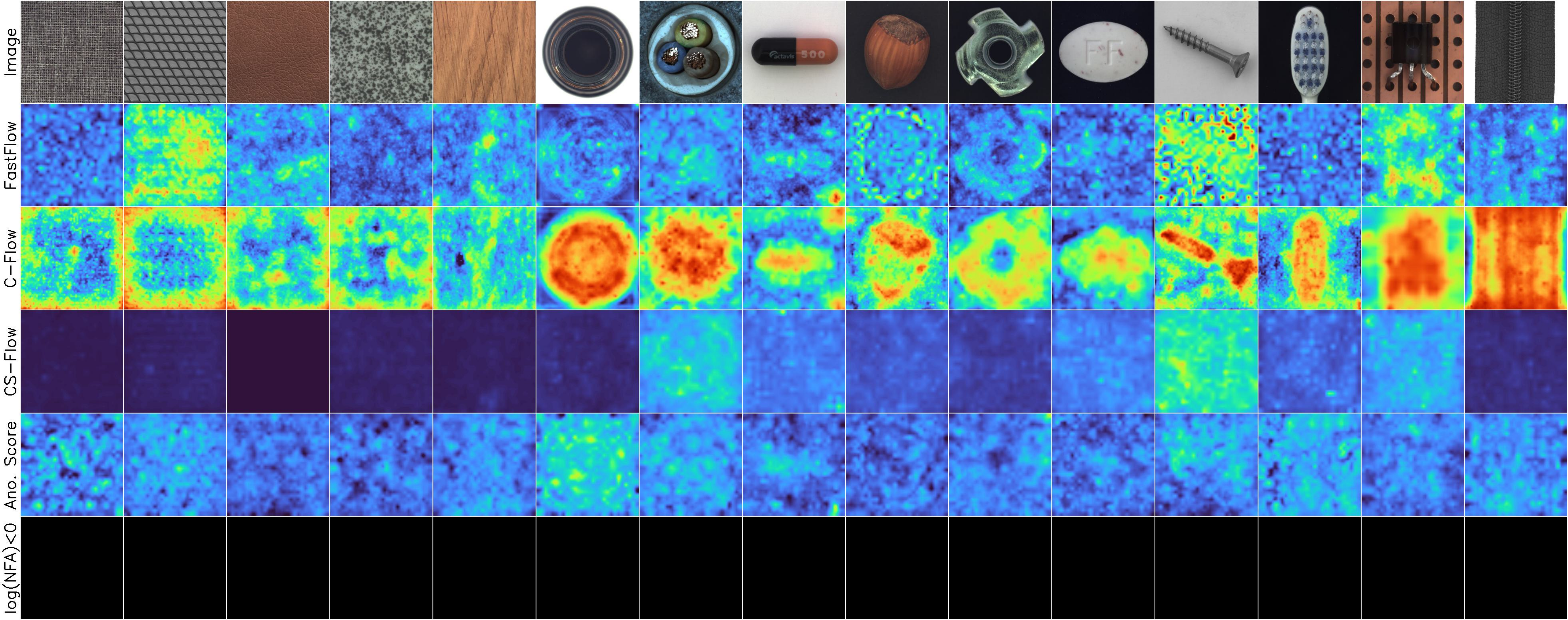}
   \caption{Normal image examples for all MVTec-AD categories. As can be seen, we always predict low values in the anomaly maps, and no detections are made.}
   \vspace{-10pt}
   \label{fig:results_good}
\end{figure*}


\section{Experimental Results}
\label{sec:experiments}

The proposed approach was tested and compared with state-of-the-art methods, by conducting extensive experimentation on several 
datasets: \textbf{MVTec-AD}~\cite{bergmann2019mvtec}, \textbf{BeanTech} (BT)~\cite{mishra2021vt-beantech}, \textbf{LGG-MRI} (MRI)~\cite{buda2019association-lggmri}, and \textbf{ShanghaiTech Campus} (STC)~\cite{liu2018ano_pred-stc}. 

\textbf{MVTec-AD} is the most widely used benchmark for anomaly detection, as most state-of-the-art anomaly detection methods report their performances on this dataset. It consists of 15 categories of textures and objects, simulating real-world industrial quality inspection scenarios with 5,000 images. Each category has normal (i.e., anomaly-free) images for training, and the testing set consists of images with different kinds of anomalies and some more normal images. \textbf{BeanTech} is also a real-world industrial anomaly dataset. It contains 2830 images of three categories, including normal and defective images. 

Besides testing the proposed method on the industrial inspection task (which is the motivation and focus of this work), we also include experimentation with data from other fields to demonstrate the generalization capability of our method. To do so, we test and compare our method on datasets from the medical and surveillance fields. \textbf{LGG-MRI} contains images of lower-grade gliomas (brain tumors), with images of the tumors' evolution over time, from 110 patients of 5 institutions, taken from The Cancer Genome Atlas. Lastly, the \textbf{ShanghaiTech Campus} dataset contains frames extracted from videos of 13 different scenes of surveillance cameras of the mentioned campus.

For assessing the anomaly maps defined in \eqref{ec:as-likelihood}, we adopt the \textit{AUROC} metric (the area under the Receiver Operating Characteristic curve) at a pixel level, as it is the most widely used metric in the literature. We also consider \textit{AUPRO} (the area under the per-region-overlap curve) since it ensures that both large and small anomalies are equally important. For assessing the anomaly masks (after thresholding), we use the \textit{mIoU} metric (Mean Intersection over Union). Finally, even though the image-level anomaly detection task is not the primary focus of this work, we have included the image-level \textit{AUROC} results in Appendix~\ref{app:imagelevel} because we understand that it is important to check that any anomaly localization method should also perform well when used to classify an entire image as anomalous or normal (the detection task).

\subsection{Results on the MVTec-AD dataset}

Before proceeding to present the results on this dataset, it is worth mentioning that since the trained models for FastFlow, CFlow, and CS-Flow are not available, it was necessary to re-train them for all the categories. To ensure a fair comparison, we performed numerous training runs until the obtained \textit{AUROC} were comparable to the ones reported in the original papers. Note that for FastFlow and CFlow, a different set of hyper-parameters is chosen for each category, sometimes varying even the number of \textit{flow steps} and the image resolution, while for our method, the architecture is always kept the same.

The results obtained for the anomaly maps assessment in terms of \textit{AUROC} and \textit{AUPRO} are shown in Table~\ref{tab:pixel_auroc} and Table~\ref{tab:pixel_aupro}, respectively. U-Flow achieves state-of-the-art results, outperforming all previous methods on average for \textit{AUROC}. Considering the \textit{AUPRO} metric, our results rank first among the flow-based methods and second overall, achieving the best results for several categories. It is worth emphasizing  that FastFlow and CFlow results are reported using different hyperparameters that even change the architecture, and this is somehow unfair in their favor. Our results are always superior to CS-Flow, which also uses a multi-scale strategy inside the NF.


In addition, besides obtaining excellent results for \textit{AUROC} and \textit{AUPRO} in the anomaly localization task, our method presents another significant advantage with respect to all others: it produces a fully unsupervised segmentation of the anomalies and significantly outperforms its competitors,
as shown in the next section. \\

For visual assessment, we include Figure~\ref{fig:results}, which presents typical example results of anomalous images from various categories and their comparison with other methods, and Figure~\ref{fig:results_good}, which shows typical examples of normal images in order to check that the anomaly maps present low values and that the detection method does not produce false positives. 

\subsubsection{Segmentation results}
\label{sec:segmentation-results}

Providing an operation point is crucial in almost any industrial application. Most recent deep learning industrial anomaly detection methods in the literature focus on generating anomaly maps and evaluating them using the \textit{AUROC} metric; they do not provide detection thresholds or anomaly segmentation masks. Some methods do provide segmentation results, but these are not consistent with a realistic scenario: they are obtained using an \textit{oracle}-like threshold that is computed by maximizing some metric over the test images~\cite{roth2022towards-patchcore,tsai2022multi-mspba,gudovskiy2022cflow}. 


In this section, we report results concerning anomaly segmentation based on the standard unsupervised threshold for meaningful events: $\text{NFA} \leq 1$ ($\log (\text{NFA}) \leq 0$). As detailed in Section~\ref{subsec:nfa}, this threshold implies that, theoretically, we allow one false anomaly detection per image on average. 
Moreover, this threshold can be tuned for any application-specific needs. For instance, in an industrial environment with the goal of detecting sample defects, an average value of one false detection per image may not be a good choice, as it would raise too many false alarms in normal images (one per image, on average). Instead, in this case, the threshold could be customized to anticipate, for example, a designated frequency of false alarms within a defined timeframe. However, as we will show next, the proposed method is inherently very robust to the threshold choice, and a much more restrictive threshold on the false positives can be considered without affecting the true positives' rate. \\ 


For evaluating the anomaly segmentation performance, we propose to use the \textit{mIoU} metric. Compared to \textit{IoU}, using the average of all images (\textit{mIoU}) prevents large anomaly regions from having more weight in the overall metric than small ones, and therefore, this metric better reflects how good the methods are in detecting all anomalies. The results for the \textit{mIoU} metric are shown in Table~\ref{tab:miou}. We limit this comparison to the flow-based methods. As the state-of-the-art methods to which we compare do not provide detection thresholds, we adopt two strategies: {\em (i)} we compute an \textit{oracle}-like threshold which maximizes the \textit{mIoU} for the testing set, and {\em (ii)} we use a \textit{fair} strategy that only uses training data to find the threshold. In the latter, the threshold is set to allow on average one false positive in the training (anomaly-free) images, as it would be analogous to setting $\text{NFA} \leq 1\text{ false alarm}$. \\

Visual examples of the obtained detections can also be found in Figure~\ref{fig:results} for anomalous images, and in Figure~\ref{fig:results_good} for normal images (which present no detections).

\begin{figure*}[t]
  \centering
  \includegraphics[width=1\linewidth]
  {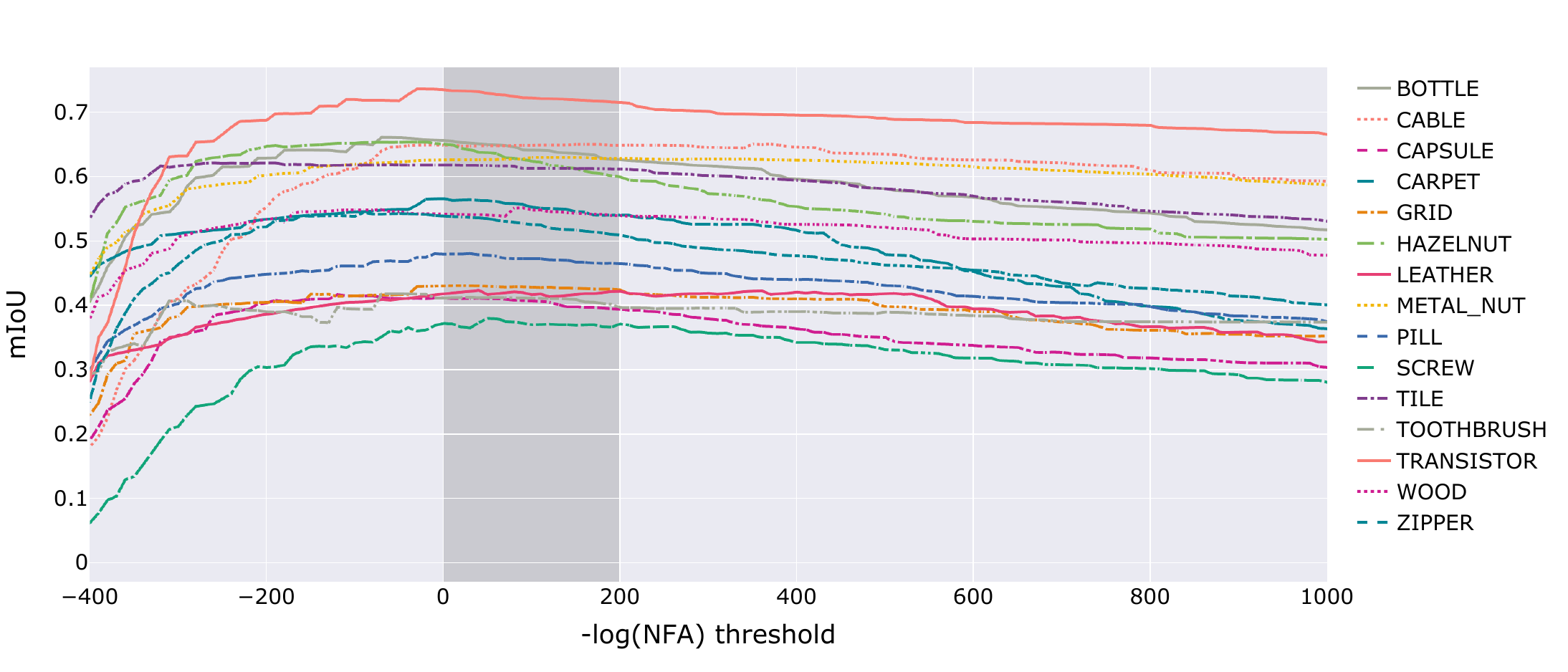}
   \caption{Segmentation results (\textit{mIoU}) as a function of the $-\log(\text{NFA})$ threshold for all MVTec-AD categories. The automatic threshold $-\log(\text{NFA}) = 0$ is always very close to the optimal threshold for each category (which corresponds to the maximum of each curve). In addition, there is a wide range of thresholds, approximately indicated with the grayed-out region in the chart, for which the performance remains almost the same, indicating that this strategy is robust and, in practice parameter-free.}
   \label{fig:mious}
\end{figure*}

As seen in Table~\ref{tab:miou}, our automatic thresholding strategy significantly outperforms all others, even when compared with their \textit{oracle}-like threshold. 
Additionally, it is important to note that our results obtained with the oracle-like and automatic thresholds are very close. This demonstrates the validity of the proposed statistical derivation and the inter-scale independence achieved by the proposed architecture. To further illustrate this point and to show the robustness of the anomaly detection method with respect to the threshold on the NFA, Figure~\ref{fig:mious} depicts the \textit{mIoU} for all the MVTec-AD categories, as a function of $-\log(\text{NFA})$ threshold. These graphs show that the unsupervised threshold  $\log(\text{NFA})\leq 0$ is always near the optimal point (the oracle threshold), corresponding to the maximum \textit{mIoU}, that is reported in Table~\ref{tab:miou} (noted as ``Oracle threshold $\xrightarrow{}$ Ours''). Not only are the oracle and the automatic thresholds close, but the variation in \textit{mIoU} is very low in a wide range of possible thresholds, showing that this detection strategy is robust and parameter-free in practice. \\

As a verification, we additionally include a numerical evaluation of the proposed method's performance under different false alarm thresholds in the Supplementary Material. When allowing one false alarm per image on average (with the threshold set as $\log(NFA)\leq 0$), the resulting average \textit{mIoU} was found to be 0.541. Similarly, when permitting one false alarm per one million images on average (threshold $\log(NFA)\leq -6$), the obtained \textit{mIoU} was almost identical, at 0.540. 
The robustness of our method to threshold variations can be attributed to its underlying approach. It utilizes a hierarchical structure based on the set of connected components derived from level sets, identifying maximal significant regions, defined as regions that potentially encompass lesser significant ones and are not encompassed by any higher significant region. Consequently, our method does not explicitly evaluate all potential regions. This characteristic contributes to computational efficiency and enhances the robustness of threshold selection. Each detected region encompasses a set of regions, and the ultimately identified regions are large deviations from the background model.
Finally, we conducted detections on all anomaly-free test images to evaluate the empirical false alarm rate. Out of 467 images analyzed, 6 exhibited false alarms, of which 4 were identified as genuine yet unlabeled anomalies. The remaining 2 instances were confirmed as true false alarms, attributable to significant deviations from the null hypothesis observed in the Normalizing Flow embedding. It is worth mentioning that these two false alarms are two sets of detections, as discussed before, suggesting that the actual average false alarm rate could be closer to the theoretical one. We refer the reader to the Supplementary Material for a more detailed analysis.

\subsection{Results on other datasets}
\label{sec:more-results}

\begin{figure*}[t]
  \centering
   \includegraphics[width=1\linewidth]
   {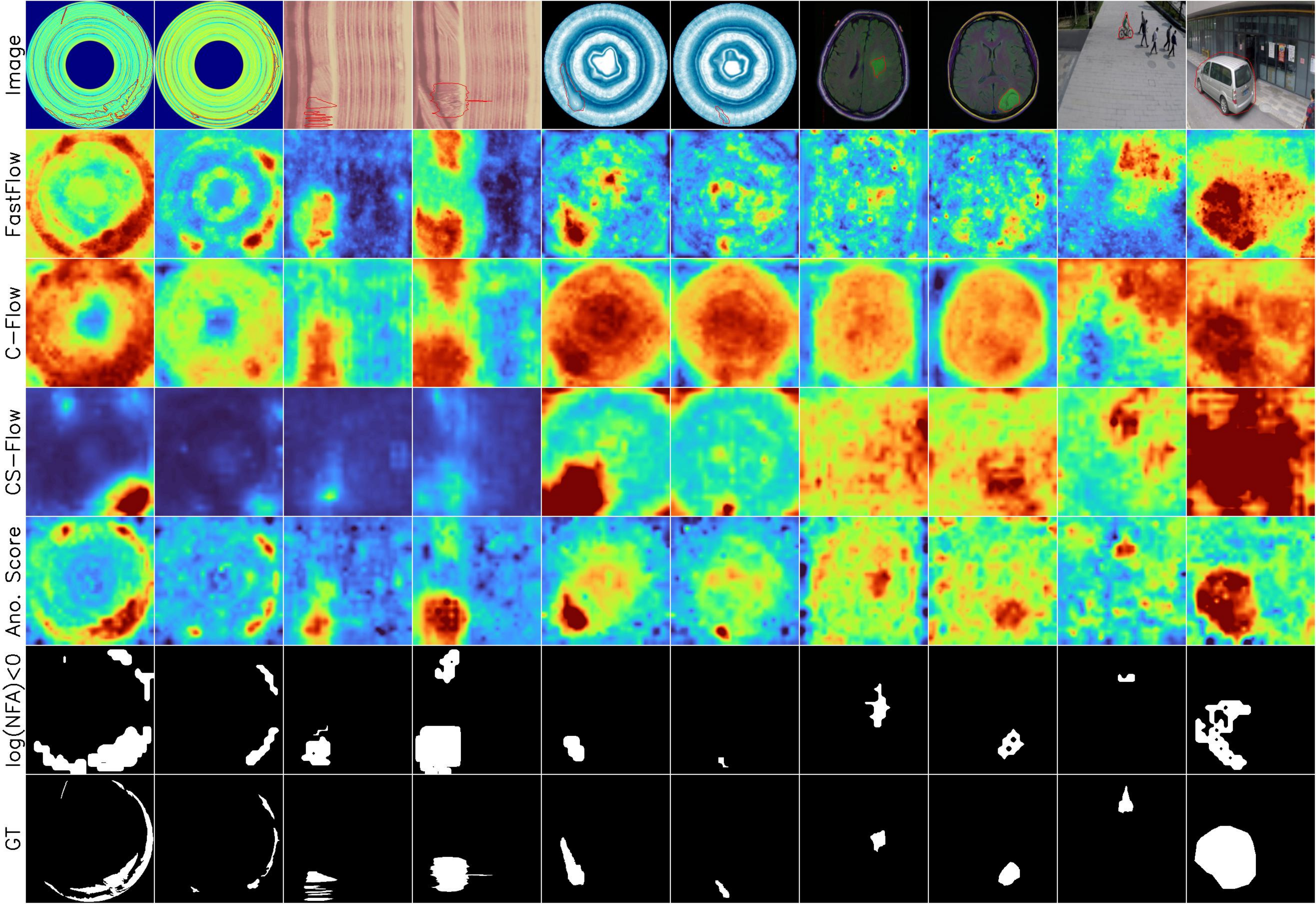}
   \caption{Examples of typical results on BeanTech, LGG MRI, and STC datasets. The first row shows the original images with over-imposed ground truth. The second, third, and fourth rows show the results of FastFlow, CFlow, and CS-Flow for comparison. The following two rows are the outputs of our method: the anomaly score defined in~\eqref{ec:as-likelihood}, and the anomaly segmentation with $\log(\text{NFA}) < 0$, and the last row is the ground truth. Again, our method achieves very good visual and numerical results and is able to detect the anomalies with great confidence. All detections of these examples exhibit very low log(NFA) values, ranging from -56 to -1586.}
   \label{fig:more_results}
\end{figure*}

In this section, we present the results obtained on the other datasets: BeanTeach (BT), LGG MRI (MRI), and ShanghaiTech Campus (STC). Although our work is motivated by the industrial inspection task, we also evaluate the performance on datasets that are very different from that scenario. The results reported here demonstrate the robustness and generalization capability of the proposed approach (c.f. Table~\ref{tab:more_results}). For all datasets, we obtained excellent results, reaching top performance for almost all metrics and datasets. 
Again, all results were obtained by training the other methods with different hyperparameters to achieve the best possible results. In cases where some combinations of metrics and datasets were reported in the original articles, we confirmed that the same results had been obtained. Typical qualitative results are shown in Figure~\ref{fig:more_results}.

\begin{table}[t]
  
  \small
  
  \centering
  \begin{tabularx}{1\columnwidth}{@{}c@{\hskip 2pt}|@{\hskip 2pt}Y@{\hskip 4pt}Y@{\hskip 4pt}Y@{\hskip 4pt}Y@{}}
    \toprule
    \begin{tabular}{@{}c@{}}\textbf{Feature}\\ \textbf{Extractor}\end{tabular} & \begin{tabular}{@{}c@{}}\textbf{Fast}\\ \textbf{Flow}\end{tabular} & \begin{tabular}{@{}c@{}}\textbf{CFlow}\\ \end{tabular}  & \begin{tabular}{@{}c@{}}\textbf{CS-}\\ \textbf{Flow}\end{tabular} & \begin{tabular}{@{}c@{}}\textbf{U-Flow}\\ \textbf{(ours)}\end{tabular}  \\
    \midrule
    ResNet18          & 4.9 M    & 5.5 M  & -      & \textbf{4.3 M}  \\
    WideResnet50      & 41.3 M   & 81.6 M & -      & \textbf{34.8 M} \\
    EfficientNet      & -        & -      & 275 M  & -               \\
    CaIT M48          & 14.8 M   & 10.5 M & -      & \textbf{8.9 M}  \\
    \textbf{MS-CaiT}  & -        & -      & -      & \textbf{12.2 M} \\
    \bottomrule
  \end{tabularx}
  \caption{Complexity analysis: comparison of the number of trainable parameters for different feature extractors and methods.}
  \label{tab:complexity}
\end{table}

\renewcommand{\b}[1]{\textbf{#1}}
\newcommand{\vtext}[1]{\begin{tabular}{@{}c@{}}#1\end{tabular}}

\begin{table*}[t]

  \small
  \centering
    
  \begin{tabularx}{1\textwidth}{ @{\hskip 8pt} c @{\hskip 8pt} | @{\hskip 10pt} l @{\hskip 8pt} YYYY Y Y}
    \toprule
    \multicolumn{1}{c}{} & & \textbf{LGG~MRI} & \textbf{STC} & \textbf{BT01} & \textbf{BT02} & \textbf{BT03} & \textbf{BT~AV.}\\
    \midrule
        
    \multirow{4}{*}{\vtext{\rotatebox[origin=c]{90}{\textbf{AUROC}}}}
        & \textbf{FastFlow}          & 82.51     & 95.51     & 95.98     & 96.69     & 99.32     & 97.33     \\
        & \textbf{CFlow}             & 86.41     & \b{96.90} & 94.29     & 96.13     & 99.58     & 96.67     \\
        & \textbf{CS-Flow}           &  81.82    & 86.68     & 91.66     & 94.89     & 99.10     & 95.22     \\
        & \textbf{U-Flow (ours)}     & \b{93.55} & 87.19     & \b{97.70} & \b{96.93} & \b{99.76} & \b{98.13} \\
    \midrule
    \multirow{4}{*}{\vtext{\rotatebox[origin=c]{90}{\textbf{AUPRO}}}}
        & \textbf{FastFlow}          & 47.48     & 73.67      & 76.23     & 62.39     & 96.60     & 78.41     \\
        & \textbf{CFlow}             & 59.47     & \b{82.05}  & 60.19     & 54.72     & 98.32     & 71.08     \\
        & \textbf{CS-Flow}           & 37.30     & 54.04      & 73.75     & 47.18     & 94.45     & 71.79     \\
        & \textbf{U-Flow (ours)}     & \b{71.74} & 47.54      & \b{80.06} & \b{66.49} & \b{98.83} & \b{81.79} \\
    \midrule
    \multirow{4}{*}{\vtext{\rotatebox[origin=c]{90}{\begin{tabular}{@{}c@{}}\textbf{mIoU}\\ \textbf{Oracle}\end{tabular}}}} 
        & \textbf{FastFlow}      & 0.2181     & 0.4840     & 0.4813     & \b{0.2929} & 0.9297     & \b{0.5680} \\
        & \textbf{CFlow}         & 0.2182     & 0.4831     & 0.4267     & 0.2782     & 0.9323     & 0.5457     \\
        & \textbf{CS-Flow}       & 0.2181     & 0.4895     & 0.3776     & 0.2439     & 0.9390     & 0.5202     \\
        & \textbf{U-Flow (ours)} & \b{0.2279} & \b{0.5052} & \b{0.4880} & 0.2620     & \b{0.9414} & 0.5638     \\
    \midrule
    \multirow{4}{*}{\vtext{\rotatebox[origin=c]{90}{\begin{tabular}{@{}c@{}}\textbf{mIoU}\\ \textbf{Fair}\end{tabular}}}} 
        & \textbf{FastFlow}      & 0.1947     & 0.4703     & 0.4408     & \b{0.2890} & 0.9129     & 0.5476     \\
        & \textbf{CFlow}         & \b{0.2175} & 0.4831     & 0.4227     & 0.2686     & 0.9149     & 0.5354     \\
        & \textbf{CS-Flow}       & 0.2060     & 0.4211     & 0.3522     & 0.1353     & 0.8738     & 0.4538     \\
        & \textbf{U-Flow (ours)} & 0.1467     & \b{0.4950} & \b{0.4667} & 0.2448     & \b{0.9330} & \b{0.5482} \\
        
    \bottomrule
  \end{tabularx}
  \caption{Results for LGG MRI~\cite{buda2019association-lggmri}, ShanghaiTech Campus (STC)~\cite{liu2018ano_pred-stc}, and BeanTeach (BT)~\cite{mishra2021vt-beantech} datasets. Comparison with best-performing flow-based models: FastFlow~\cite{yu2021fastflow}, CFlow~\cite{gudovskiy2022cflow}, and CS-Flow~\cite{csflow-rudolph2022fully}. Our method obtains the best results for almost all metrics and datasets, outperforming the other methods. 
  Both \textit{AUROC} and \textit{AUPRO} refer to the pixel-level metric (localization task). For \textit{mIoU}, we present the results using both the ``Fair'' and the ``Oracle''-like thresholds.}
  \label{tab:more_results}
\end{table*}

\subsection{Implementation details and complexity}
\label{subsec:details}

The method was implemented in PyTorch~\cite{NEURIPS2019_9015-pytorch}, using PyTorch Lightning~\cite{Falcon_PyTorch_Lightning_2019}. The NFs were implemented using the FrEIA Framework~\cite{freia}, and for all tested feature extractors, we used PyTorch Image Models~\cite{rw2019timm}. In all cases, training was performed in a \textit{GeForce RTX 2080 Ti}. 

For the MS-CaIT, the input sizes are 448x448 and 224x224 pixels. The Normalizing Flow has 2 \textit{flow stages} with 4 \textit{flow steps} each.

\label{sec:complexity}

As mentioned, our method only uses four \textit{flow steps} in each scale. As a result, it has fewer trainable parameters than FastFlow and CFlow for the same feature extractors, as shown in Table~\ref{tab:complexity}.

Although all the results reported in this section were obtained using MS-CaiT as a feature extractor, we stress that the method and the code are developed agnostic to the feature extractor and can be easily changed according to the user's preference. In the following section, we present some results testing different feature extractors. In those cases, the number of scales and the volumes' sizes could vary depending on the chosen feature extractor.

\section{Ablation Study}
\label{sec:ablation}


\begin{table*}[t]

  \small

  \centering
  \begin{tabularx}{1\textwidth}{@{}c@{\hskip 2pt} | @{\hskip 2pt} YYYYYYYYYYYYYYYr@{}}
\toprule

& Carp. & Grid & Leat. & Tile & Wood & Bott. & Cable & Caps. & HNut & MNut & Pill & Screw & Toot. & Tran. & Zipp. & Total \\

\midrule
Scale 1 & 99.08 & 97.40 & 99.32 & 94.97 & 93.45 & 98.33 & 98.11 & 98.87 & 99.09 & 97.89 & 98.67 & 99.44 & 98.74 & 96.46 & 98.55 & 97.89 \\
Scale 2 & 99.12 & 97.09 & 99.42 & 96.47 & 96.26 & 97.23 & 97.60 & 98.07 & 98.63 & 97.86 & 98.62 & 99.18 & 97.86 & 97.21 & 97.29 & 97.86 \\
Avg.   & \textbf{99.44} & 98.25 & 99.52 & 97.27 & 96.40 & 98.61 & 98.50 & 98.85 & 99.16 & 98.29 & 99.12 & \textbf{99.50} & 98.78 & 97.66 & \textbf{98.69} & 98.54 \\

\midrule
Ours    & 99.42 & \textbf{98.49} & \textbf{99.59} & \textbf{97.54} & \textbf{97.49} & \textbf{98.65} & \textbf{98.61} & \textbf{99.02} & \textbf{99.30} & \textbf{98.82} & \textbf{99.35} & 99.49 & \textbf{98.79} & 97.87 & 98.60 & \textbf{98.74} \\
\midrule

ResNet & 98.80 & 98.26 & 99.37 & 94.53 & 94.50 & 98.00 & 96.96 & 98.46 & 98.63 & 96.70 & 97.45 & 98.01 & 98.20 & \textbf{98.38} & 97.43 & 97.58 \\
Arch. & wide & r18  & r18  & wide & r18  & r18  & wide & wide & wide & wide & wide & wide & r18  & wide & wide & -$\,\,\,\,\,$ \\         
F. steps & 6 & 6 & 4 & 8 & 6 & 4 & 4 & 4 & 4 & 4 & 4 & 6 & 6 & 4 & 4 & -$\,\,\,\,\,$ \\

\midrule
MViT2 & {98.74} & {97.73} & {99.20} & {93.83} & {94.75} &  {92.79} &  {96.98} &  {98.83} &  {97.50} &  {96.39} &  {97.36} &  {97.69} &  {87.14} &  {97.00} &  {96.39} &   {96.15} \\

\bottomrule

  \end{tabularx}
  \caption{Ablation results. The {\bf middle row} (in-between horizontal lines), which shows the pixel-level \textit{AUROC} results of our proposed method (U-Flow), serves for comparison for the top and bottom halves of the table. \textbf{Top half:} ablation study for the scale-merging strategy. 
  Results using the anomaly scores generated by each scale independently, and a naive way of merging them (average). 
  \textbf{Bottom half:} ablation study for the feature extractor. The proposed method (that uses MS-CaIT) is compared with ResNet and MViT2 feature extractors. For each category we show the \textit{AUROC} of the best variant we could obtain, varying several hyper-parameters, such as the architecture (wide-resnet-50 or resnet-18) and the number of \textit{flow steps}.}
  \label{tab:ablation}
\end{table*}

In this section, we study and provide unbiased assessments regarding some of the contributions of this work: the significance of the U-shaped architecture and the benefits of the multi-scale Transformer feature extractor MS-CaiT. Both results are shown all together in Table~\ref{tab:ablation}, and explained in the following sections.

\subsection{Ablation: U-shape}
\label{sec:ablation_u}

One of the contributions of the presented work is proposing a multi-level integration mechanism by introducing the well-known U-shape architecture design to the Normalizing Flow's framework. To demonstrate that this architecture better integrates the information of the different scales, we compare the results obtained in terms of \textit{AUROC} against a modification of the architecture, in which each \textit{flow stage} runs in parallel, and the per-scale anomaly maps are merged at the end by just averaging them, as done by other methods. Additionally, we compare the results obtained using each scale separately. The results, presented in the top half of Table~\ref{tab:ablation}, show that the U-merging strategy improves the performance in almost all cases. Furthermore, this strategy, where the output of one scale inputs the next one, allows the use of less \textit{flow steps} for each scale, resulting in a network with fewer parameters, as shown in Section~\ref{sec:complexity}. 

\subsection{Ablation: feature extraction with MS-CaiT}
\label{sec:ablation_fe}

From the upper half of Table~\ref{tab:ablation}, it is also clear that utilizing image Transformers at various scales in MS-CaiT significantly improves the results, even if each of them already provides a multi-scale representation. Note that both scale-merging strategies outperform the results obtained by each individual Transformer.

In addition, in this section, we compare the results obtained by the proposed MS-CaIT feature extractor with the most common ResNet variants used in the literature: \textit{ResNet-18} and \textit{Wide-ResNet-50}, and the multi-scale Transformer MViT2. For all feature extractors, we extract features from all scales and proceed in the exact same way as presented before. Table~\ref{tab:ablation}'s bottom half makes it clear that the MS-CaiT feature extractor performs far better than ResNet variants and MViT2. Note that, in favor of the ResNet extractors, for each category, we picked the best result obtained by both variants and also varied several hyper-parameters, for example, the amount of \textit{flow steps} in each scale, while for MS-CaIT we always use the same exact architecture. Finally, note that our feature extractor always achieves better performance compared to MViT2, which is also a Transformer with a multi-scale hierarchy, probably getting some benefit from the training independence for each scale. The hyperparameter search for ResNet and MViT2 variants was performed with Optuna~\cite{optuna_2019}, using the TPE~\cite{optuna_tpesampler} sampler.

\section{Conclusion}
\label{sec:conclusion}

In this work, we introduced a novel anomaly detection method that achieves state-of-the-art results on various datasets and even outperforms them in most cases. By using modern techniques with outstanding performance, such as Transformers and Normalizing Flows, we developed a method that exploits their characteristics to integrate them with classic statistical modeling. Our approach consists of four phases that follow one another, and we presented a clear and compelling contribution for each one:
{\em (i)} We propose a new feature extractor using pre-trained Transformers to build a multi-scale representation. {\em (ii)} We integrate the U-shape architecture into the Normalizing Flow framework, creating a complete invertible architecture that settles the theoretical foundations for the NFA computation by ensuring independence in the embedding not only intra-scale but also inter-scales. {\em (iii)} We compute a likelihood-based anomaly score for each pixel with state-of-the-art performance in various datasets. And {\em (iv)} we propose a new anomaly detection method that permits to derive an unsupervised threshold based on the \textit{a contrario} framework; this method exploits the statistical independence of the U-Flow embeddings to build a background model that produces excellent anomaly segmentation results. 

The proposed approach was extensively evaluated and compared with the top-performing methods in the literature using different metrics, obtaining state-of-the-art results and, in most cases, even outperforming all previous methods. Finally, the method was also applied to several datasets from different domains with excellent results, demonstrating a strong generalization capability.

\section*{Acknowledgments}
The authors thank Agencia Nacional de Investigación e Innovación, Uruguay, for partially funding this work through a graduate scholarship.


\begin{table*}[t]
  
  \small
  
  \centering
  \begin{tabularx}{1\textwidth}{@{}c@{}YYYYYYYYY@{}}
\toprule


\textbf{Category} & \begin{tabular}{@{}c@{}}\textbf{P.SVDD} \\  \end{tabular} & \begin{tabular}{@{}c@{}}\textbf{SPADE}\\ \end{tabular} & \begin{tabular}{@{}c@{}}\textbf{Cut-} \\ \textbf{Paste} \end{tabular} & \begin{tabular}{@{}c@{}}\textbf{Patch} \\ \textbf{Core} \end{tabular}& \begin{tabular}{@{}c@{}}\textbf{PEFM}\\  \end{tabular} & \begin{tabular}{@{}c@{}}\textbf{Fast} \\ \textbf{Flow} \end{tabular} & \begin{tabular}{@{}c@{}}\textbf{CFlow}\\  \end{tabular} & \begin{tabular}{@{}c@{}}\textbf{CS-Flow}\\  \end{tabular} & \begin{tabular}{@{}c@{}}\textbf{U-Flow}\\ \textbf{(ours)}\end{tabular} \\


\midrule
\textbf{Carpet}      &  92.90            &  98.60  &  \textbf{100.0}  &  98.70           &  \textbf{100.0}  &  \textbf{100.0} &  \textbf{100.0}  & \textbf{100.0} & \textbf{100.0} \\
\textbf{Grid}        &  94.60            &  99.00  &  96.20           &  98.20           &  96.57           &  99.70          &  97.60           & 99.00          & \textbf{99.75} \\
\textbf{Leather}     &  90.90            &  99.50  &  95.40           &  \textbf{100.0}  &  \textbf{100.0}  &  \textbf{100.0} &  97.70           & \textbf{100.0} & \textbf{100.0} \\
\textbf{Tile}        &  97.80            &  89.80  &  \textbf{100.0}  &  98.70           &  99.49           &  \textbf{100.0} &  98.70           & \textbf{100.0} & \textbf{100.0} \\
\textbf{Wood}        &  96.50            &  95.80  &  99.10           &  99.20           &  99.19           &  \textbf{100.0} &  99.60           & \textbf{100.0} & 99.91          \\
\midrule
\textbf{Av. texture} &  94.54            &  96.54  &  98.14           &  98.96           &  99.05           &  \textbf{99.94} &  98.72           & 99.80 & 99.93          \\
\midrule
\textbf{Bottle}      &  98.60            &  98.10  &  99.90           &  \textbf{100.0}  &  \textbf{100.0}  &  \textbf{100.0} &  \textbf{100.0}  & 99.80 & \textbf{100.0} \\
\textbf{Cable}       &  90.30            &  93.20  &  \textbf{100.0}  &  99.50           &  98.95           &  \textbf{100.0} &  \textbf{100.0}  & 99.10 & 98.97          \\
\textbf{Capsule}     &  76.70            &  98.60  &  98.60           &  98.10           &  91.90           &  \textbf{100.0} &  99.30           & 97.10 & 99.56          \\
\textbf{Hazelnut}    &  92.00            &  98.90  &  93.30           &  \textbf{100.0}  &  99.89           &  \textbf{100.0} &  96.80           & 99.60 & 99.71          \\
\textbf{Metal nut}   &  94.00            &  96.90  &  86.60           &  \textbf{100.0}  &  99.85           &  \textbf{100.0} &  91.90           & 99.10 & \textbf{100.0} \\
\textbf{Pill}        &  86.10            &  96.50  &  99.80           &  96.60           &  97.51           &  99.40          &  \textbf{99.90}  & 98.60 & 98.80          \\
\textbf{Screw}       &  81.30            &  99.50  &  90.70           &  98.10           &  96.43           &  97.80          &  \textbf{99.70}  & 97.60 & 96.31          \\
\textbf{Toothbrush}  &  \textbf{100.0}   &  98.90  &  97.50           &  \textbf{100.0}  &  96.38           &  94.40          &  95.20           & 91.90 & 91.39          \\
\textbf{Transistor}  &  91.50            &  81.00  &  99.80           &  \textbf{100.0}  &  97.83           &  99.80          &  99.10           & 99.30 & 99.92          \\
\textbf{Zipper}      &  97.90            &  98.80  &  \textbf{99.90}  &  98.80           &  98.03           &  99.50          &  98.50           & 99.70 & 98.74          \\
\midrule
\textbf{Av. objects} &  90.84            &  96.04  &  96.61           &  \textbf{99.11}  &  97.68           &  99.09          &  98.04           & 98.18 & 98.34          \\
\midrule
\midrule
\textbf{Av. total}   &  92.07            &  96.21  &  97.12           &  99.06           &  98.13           &  \textbf{99.37} &  98.27           & 98.72 & 98.87          \\
\bottomrule
  \end{tabularx}
  \caption{MVTec-AD results: image-level \textit{AUROC}. Comparison with best perfroming and flow-based methods: P.SVDD~\cite{yi2020patch-svdd}, SPADE~\cite{cohen2020sub-spade}, Cut-Paste~\cite{li2021cutpaste}, PatchCore~\cite{roth2022towards-patchcore}, PEFM~\cite{wan2022position-pefm}, FastFlow~\cite{yu2021fastflow}, CFlow~\cite{gudovskiy2022cflow}, and CS-Flow~\cite{csflow-rudolph2022fully}. Our method achieves state-of-the-art results also for the image-level metric (image-level \textit{AUROC}). Among the flow-based methods, we rank second, only after FastFlow, which uses an unfair test-set tuning.}
  \label{tab:image_auroc}
\end{table*}

\renewcommand{\b}[1]{\textbf{#1}}

\begin{table*}[t]
  \small
  \centering
  \begin{tabularx}{1\textwidth}{ @{\hskip 8pt} c @{\hskip 8pt} | @{\hskip 10pt} l @{\hskip 8pt} YYYY Y Y}
    \toprule
    \multicolumn{1}{c}{} & & \textbf{LGG~MRI} & \textbf{STC} & \textbf{BT01} & \textbf{BT02} & \textbf{BT03} & \textbf{BT~AV.}\\
    \midrule
    \multirow{4}{*}{\vtext{\rotatebox[origin=c]{90}{\textbf{IMAGE}}}} & 
        \textbf{FastFlow}            & 61.49     & 74.04      & \b{100.0} & \b{89.13} & 96.65     & 95.26     \\
        & \textbf{CFlow}             & 42.56     & 66.39      & 94.95     & 79.68     & \b{99.96} & 91.53     \\
        & \textbf{CS-Flow}           & 67.41     & \b{98.46}  & 99.81     & 87.26     & 99.94     & 95.67     \\
        & \textbf{U-Flow (ours)}     & \b{72.60} & 64.59      & 99.42     & 88.72     & 99.72     & \b{95.95} \\        
    \bottomrule
  \end{tabularx}
  \caption{Image-level \textit{AUROC} results for LGG MRI~\cite{buda2019association-lggmri}, ShanghaiTech Campus (STC)~\cite{liu2018ano_pred-stc}, and BeanTeach (BT)~\cite{mishra2021vt-beantech} datasets. Comparison with best-performing flow-based models: FastFlow~\cite{yu2021fastflow}, CFlow~\cite{gudovskiy2022cflow}, and CS-Flow~\cite{csflow-rudolph2022fully}.}
  \label{tab:more_results_image}
\end{table*}

\begin{appendices}

\section{Image-level results}
\label{app:imagelevel}

As mentioned in Section~\ref{sec:experiments}, this work focuses on the anomaly localization task. Nevertheless, we understand it is important to check that the proposed method also behaves well on the image-level detection task. Therefore, we present in Table~\ref{tab:image_auroc} the results for the image-level \textit{AUROC} (detection task). To compute an image-level score, we simply used the maximum value of the anomaly score at the pixel level, defined in Section~\ref{subsec:phase3-anomaly-score} (Eq.~\eqref{ec:as-likelihood}). This simple strategy already achieves very good results: compared with all flow-based methods, our method ranks second, only behind FastFlow. However, it is worth noting that FastFlow actually uses different hyperparamenters for each category, even changing the architecture, which is an unfair test-set tuning. 

For completeness, in Table~\ref{tab:more_results_image} we also include the image-level results for all other tested datasets. Our method ranks first on average for LGG MRI and BeanTeach. 

\end{appendices}


\bibliography{sn-bibliography}

\end{document}


\title[]{Supplementary material for U-Flow:~A~U-shaped~Normalizing~Flow for Anomaly Detection with Unsupervised Threshold}

\author[]{\fnm{Matías} \sur{Tailanian}}
\author[]{\fnm{Álvaro} \sur{Pardo}}
\author[]{\fnm{Pablo} \sur{Musé}}

\maketitle

We present here a further analysis of the proposed method and more experimental results, including failure cases and limitations.

\section{Analysis of the Normalizing Flow embedding}

\begin{figure*}[ht]
  \centering
  \includegraphics[width=1\linewidth]{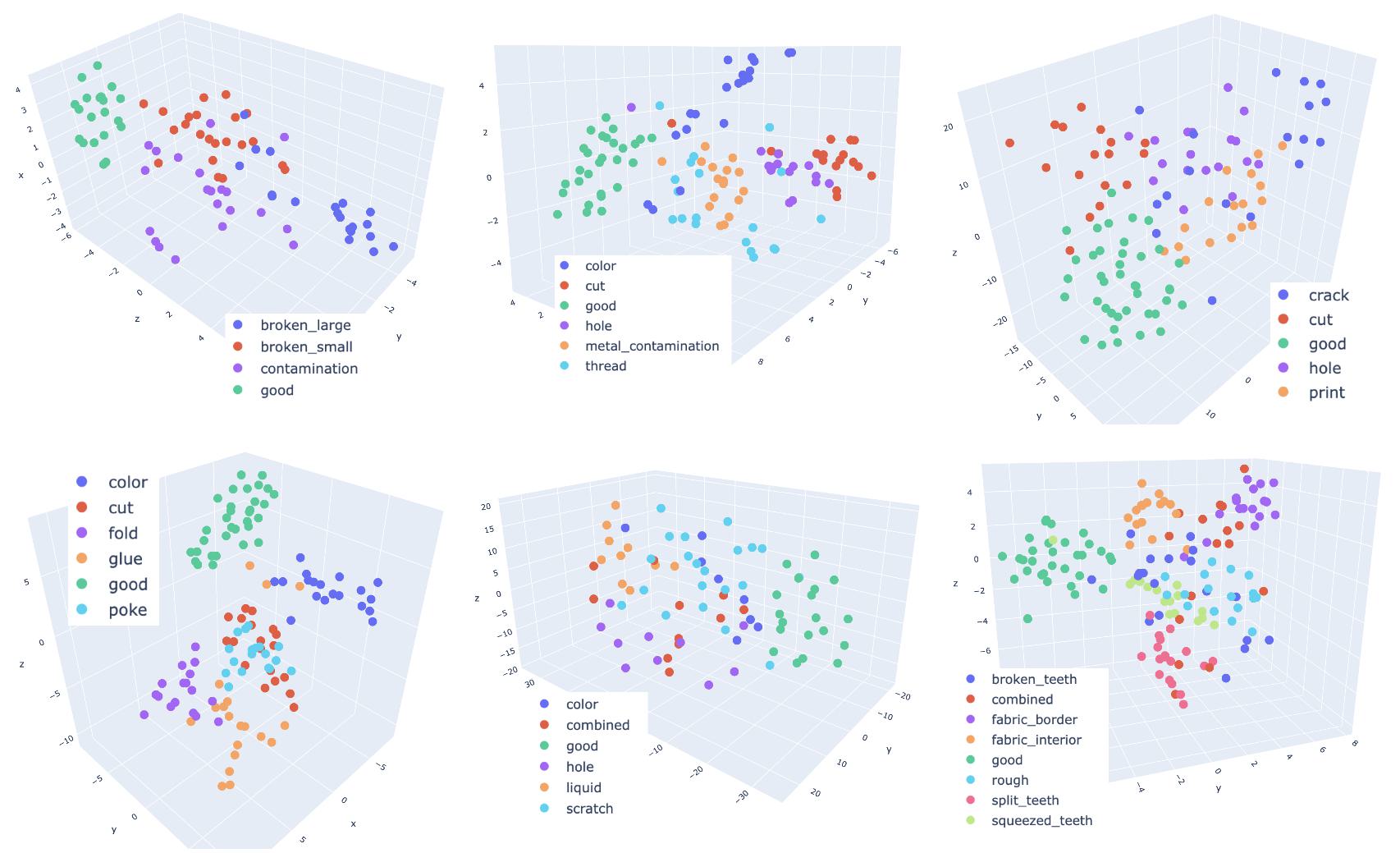}
   \caption{T-SNE embeddings. Top row: bottle, carpet and hazelnut. Bottom row: leather, wood and zipper. Normal samples are always in green, and other colors correspond to different defect types. It can be seen as a very good separation not only for the normal samples but also for each different type of defect.}
   \label{fig:embedding_sup}
\end{figure*}

To complement the results included in the main manuscript, in this section, we propose to analyze visually the embeddings produced by the NF, by mapping them to a three-dimensional space using the \textit{T-distributed Stochastic Neighbor Embedding} (T-SNE) algorithm. Results for various categories are shown in Figure~\ref{fig:embedding_sup}. Each dot in this 3D space represents a different test image, and they are colored according to the type of defect. Green dots are always normal samples (i.e. anomaly-free images).

We could apply T-SNE directly to the entire volume $z$ to generate this mapping. But instead, aiming to reduce the input dimensionality, we apply an intermediate step. Recalling that for each scale $l$, we have an embedding $z_l$ of size $[C_l, H_l, W_l]$, we keep for each $k=\{1, \dots, C_l\}$ the mean and standard deviation of the squared values of the volume, constructing a feature vector of dimension $2 * C_l$ for each image. The final feature vector, concatenating vectors from all scales, has size $2\sum_l C_l$. These feature vectors are used as input for the T-SNE algorithm. 

It is easy to observe that both normal samples and defect types reveal a clustering structure. Even though the method was aimed at separating abnormal samples from normal samples (green dots), it is interesting to note that this representation evidences that there is also a good separation between different types of defects, giving a hint that this technique could also be potentially applied for defects classification. 

\section{More example results}
\subsection{Anomalous images}
\begin{figure*}[hb]
  \centering
  \includegraphics[width=1\linewidth]{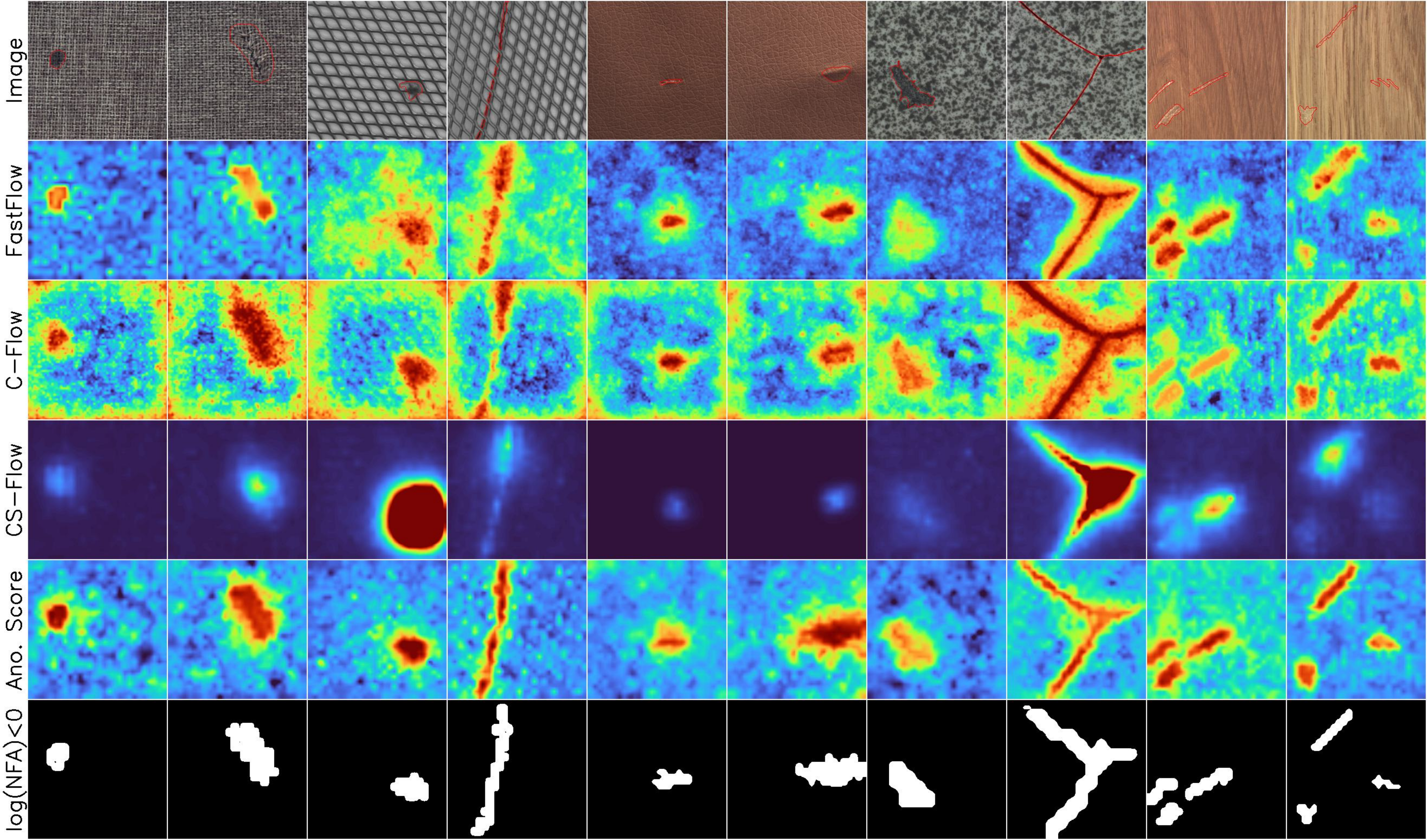}
   \caption{Textures: carpet, grid, leather, tile, and wood.}
   \label{fig:results_texture_1}
   \vspace{-1pt}
\end{figure*}

\begin{figure*}[ht]
  \centering
  \includegraphics[width=1\linewidth]{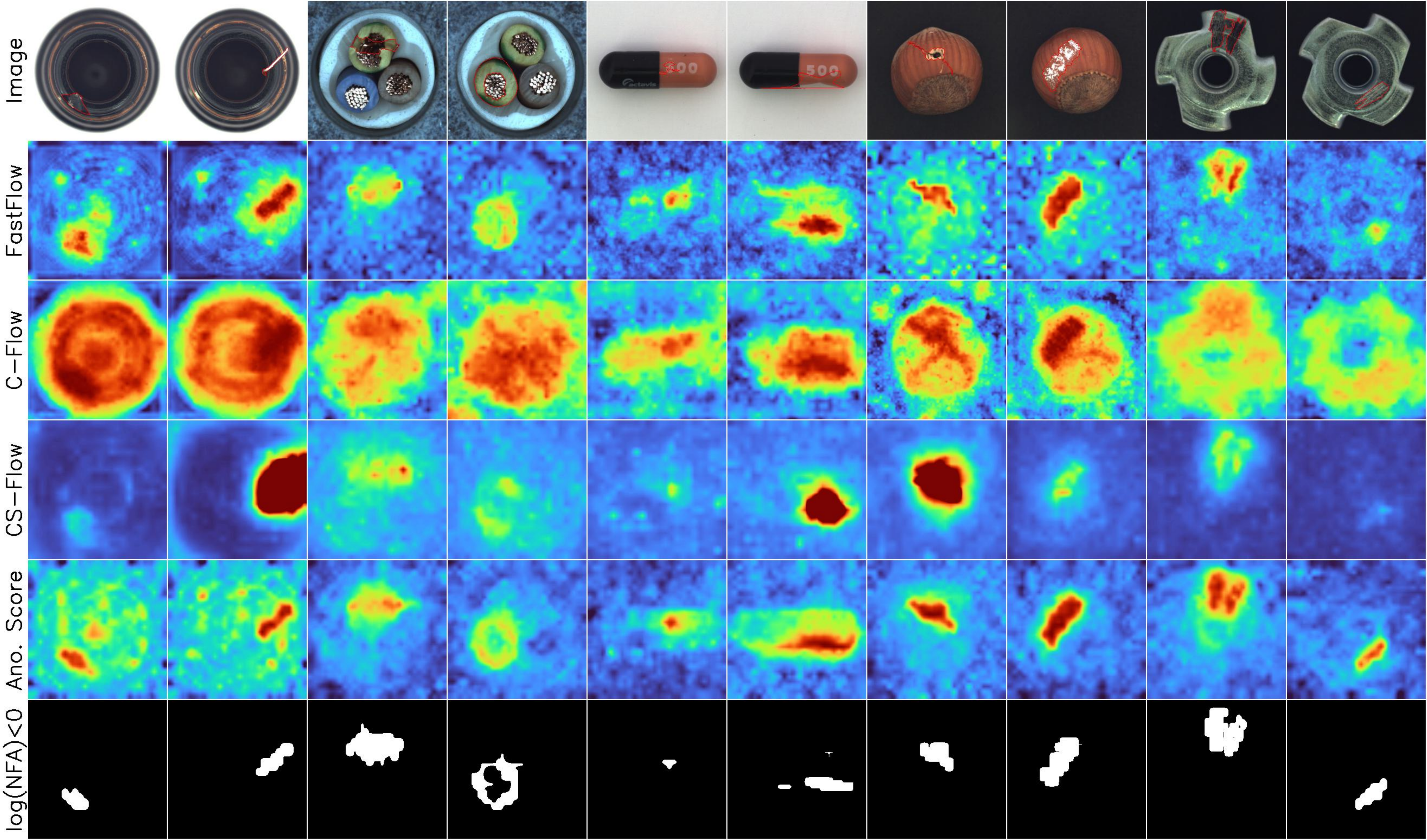}
   \caption{Objects 1: bottle, cable, capsule, hazelnut, and metal nut.}
   \label{fig:results_objects_1}
   \vspace{-10pt}
\end{figure*}

\begin{figure*}[ht]
  \centering
  \includegraphics[width=1\linewidth]{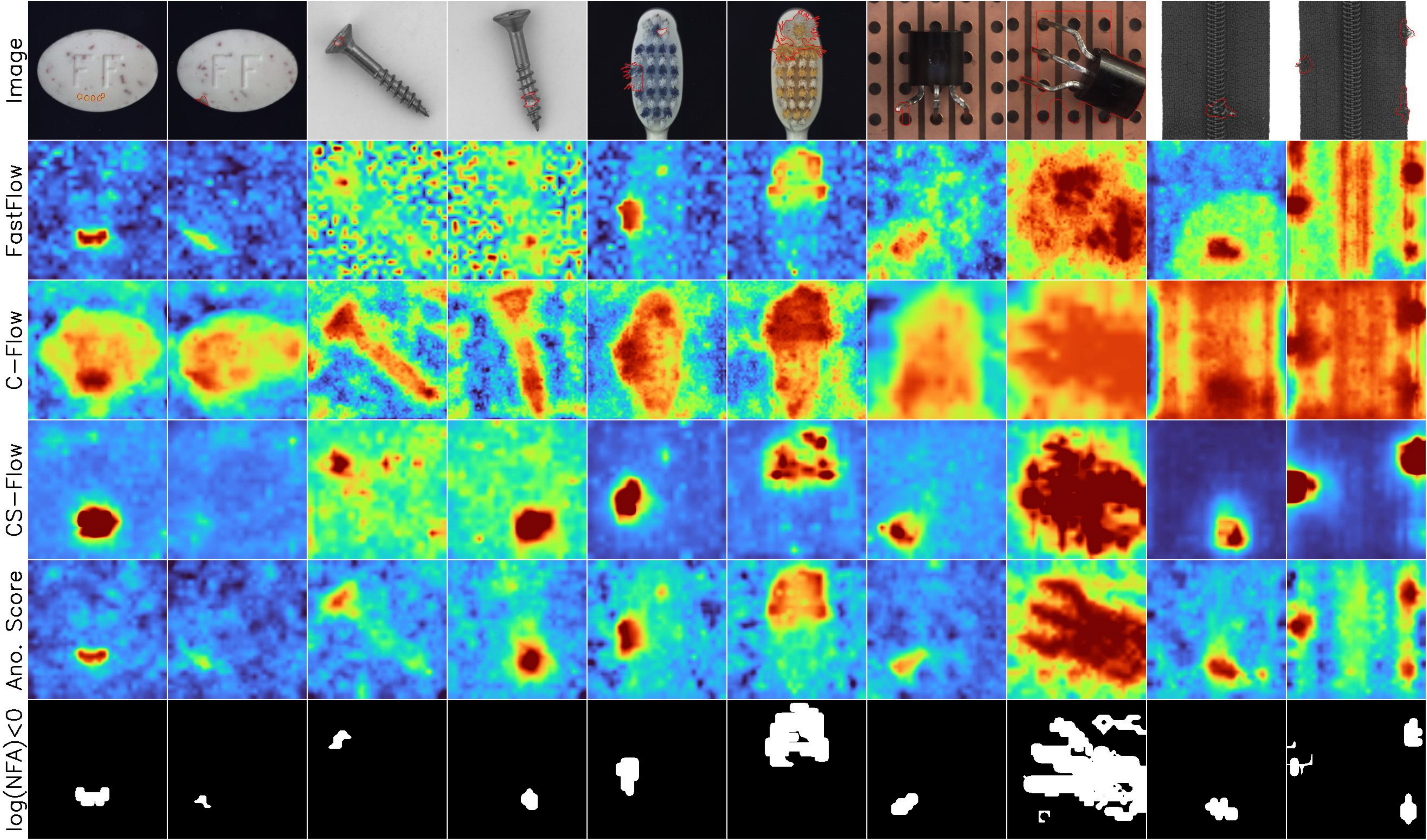}
   \caption{Objects 2: pill, screw, toothbrush, transistor, and zipper.}
   \label{fig:results_objects_2}
\end{figure*}

In this section, we present a larger set of experimental results. We display two images for each category, with different types of defects. Texture examples are shown in Figure~\ref{fig:results_texture_1}, and object examples are shown in Figures~\ref{fig:results_objects_1} and~\ref{fig:results_objects_2}.

\subsection{Normal images}

\begin{figure*}[ht]
  \centering
  \includegraphics[width=1\linewidth]{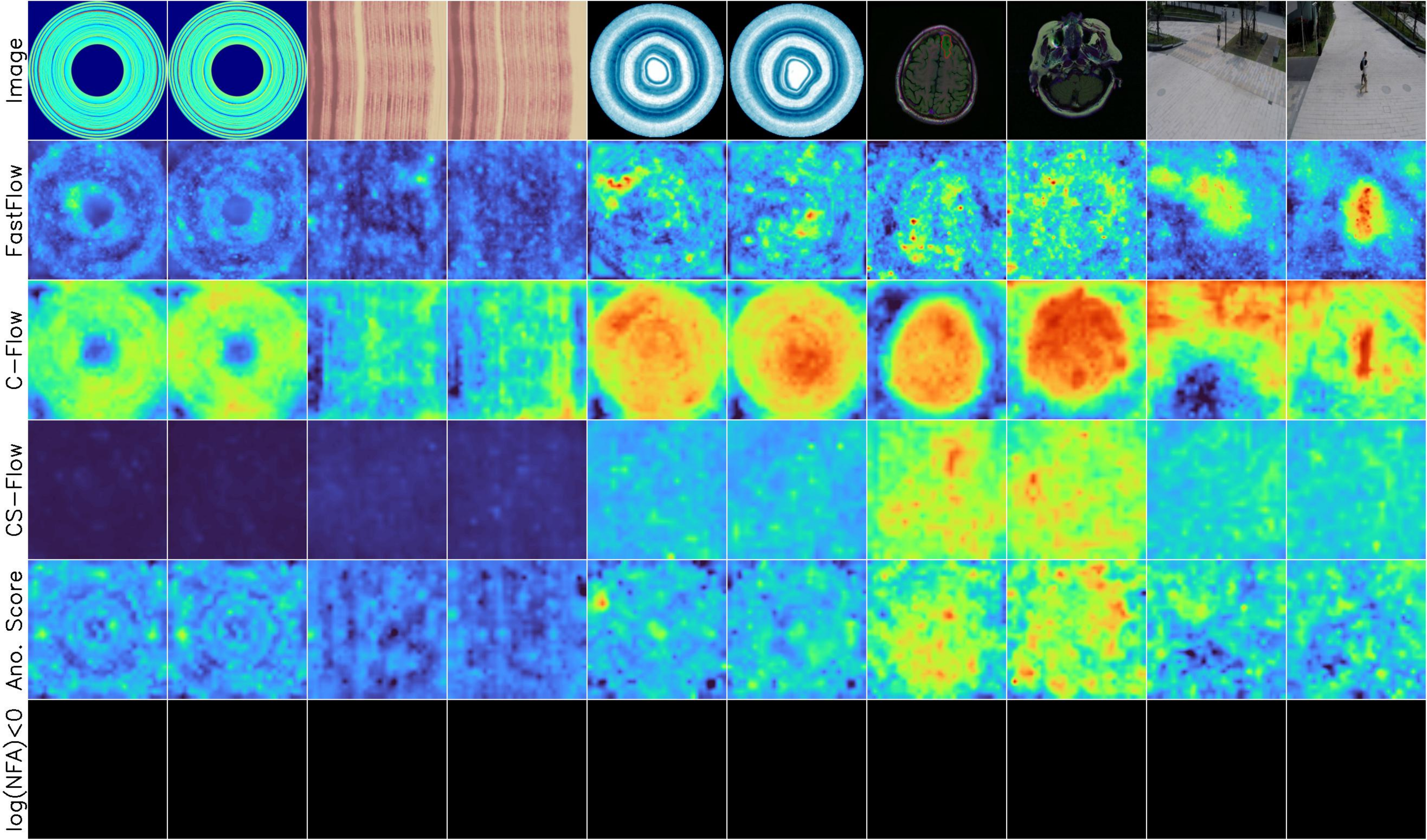}
   \caption{Normal images examples of BeanTech, LGG~MRI, and STC.}
   \label{fig:more_results_good}
\end{figure*}

Also, it is interesting to check how the anomaly scores behave with anomaly-free images, as shown in Figure~6 in the paper. In this section, we add more normal examples corresponding to the other datasets: BeanTech, LGG~MRI, and STC, in Figure~\ref{fig:more_results_good}. As can be seen, the anomaly scores are always low, and we do not detect any anomalous pixels in the segmentations.

\subsection{Unexpected surprises: detection of actual unlabeled anomalies in images}

While visually inspecting the results, we found some cases where unlabeled anomalies were detected. A careful look reveals that these structures are actually different kinds of anomalies, and therefore it is actually correct to detect them. This kind of ``false true detections'' unfairly penalizes the method's performance.

\begin{figure*}
  \centering
  \begin{subfigure}{0.8\linewidth}
    \includegraphics[width=0.99\linewidth]{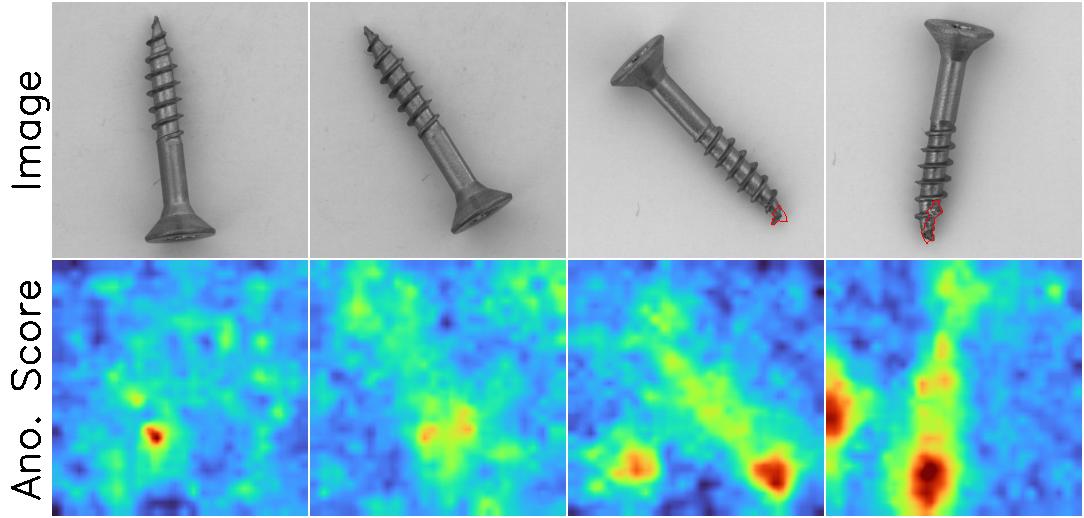}
    \caption{Example of a subtle fluff in the background for the screw category.}
    \label{fig:rare_screw}
  \end{subfigure}
  \\
  \begin{subfigure}{0.8\linewidth}
    \includegraphics[width=0.99\linewidth]{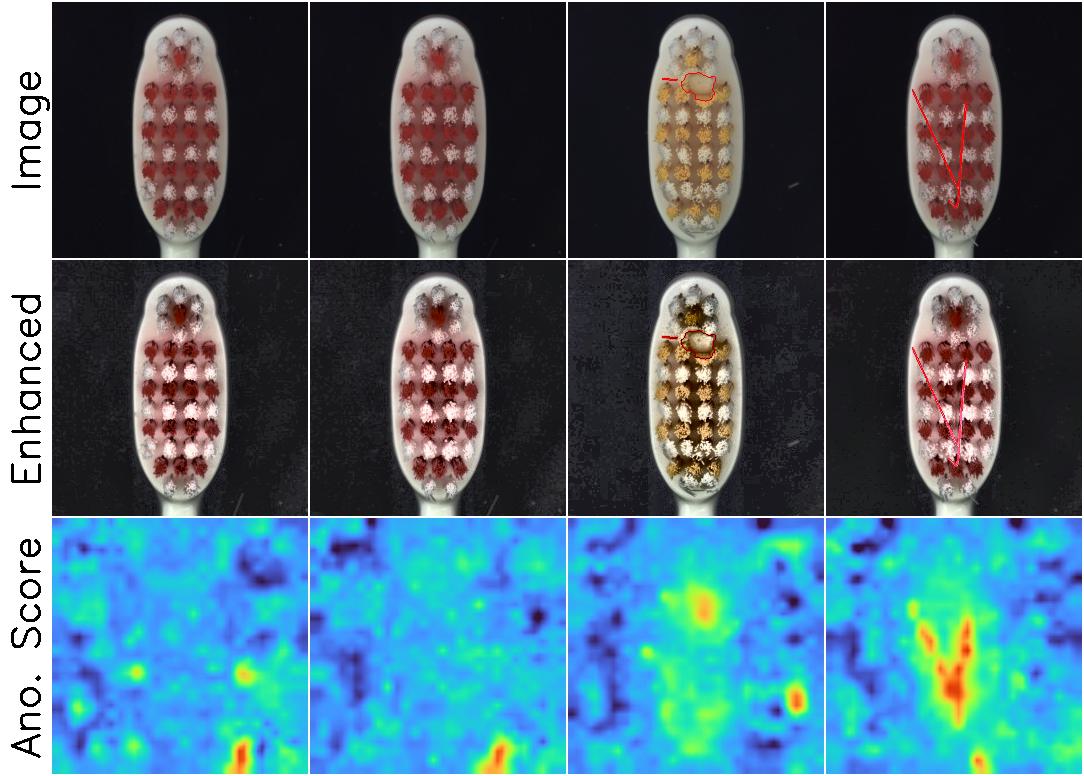}
    \caption{Examples for the toothbrush category. Some defect in the background is being detected. We include the intermediate row with a contrast-enhanced version of the image, to better identify the defect.}
    \label{fig:rare_tooth}
  \end{subfigure}
  \caption{Examples of ``false true'' anomalies. Our method detects some anomalies that are not supposed to be there, and that are not labelled. For better visualization, we recommend to zoom-in.}
  \label{fig:rare_cases}
\end{figure*}

Two such examples are shown in Figure~\ref{fig:rare_cases}. Figure~\ref{fig:rare_screw} displays four images of the screw category, in which we can observe a subtle fluff in the background, almost unnoticeable to the naked eye, that correctly stands out in the anomaly score map. The two leftmost images correspond to images labeled as anomaly-free samples, while the two rightmost ones present labeled defects somewhere else that are also correctly detected. Similarly, Figure~\ref{fig:rare_tooth} shows four images of the toothbrush category that also present some defects in the background. As they are very difficult to see with the naked eye, we included a contrast-enhanced version of the image in the middle row for visualization purposes. Again, these defects are correctly detected, but they were not supposed to be there, and since they are not labeled, they penalize the results when computing the evaluation metrics.

\begin{table*}[t]
  \small
  \centering
  \begin{tabularx}{1\textwidth}{@{}c@{\hskip 2pt} @{\hskip 2pt} YYYYYYYYYYYYYYYr@{}}
\toprule
NFA Thr. & Carp. & Grid & Leat. & Tile & Wood & Bott. & Cable & Caps. & HNut & MNut & Pill & Screw & Toot. & Tran. & Zipp. & \textbf{Avg.} \\
\midrule

1       & 0.566 & 0.431 & 0.418 & 0.619 & 0.543 & 0.656 & 0.649 & 0.412 & 0.651 & 0.627 & 0.480 & 0.373 & 0.412 & 0.735 & 0.539 & \textbf{0.541} \\
1/10      & 0.566 & 0.431 & 0.419 & 0.619 & 0.542 & 0.656 & 0.649 & 0.411 & 0.651 & 0.627 & 0.480 & 0.373 & 0.412 & 0.735 & 0.539 & \textbf{0.541} \\
1/100     & 0.566 & 0.431 & 0.419 & 0.619 & 0.542 & 0.656 & 0.649 & 0.411 & 0.651 & 0.627 & 0.480 & 0.373 & 0.412 & 0.735 & 0.539 & \textbf{0.541} \\
1/1000    & 0.566 & 0.431 & 0.419 & 0.619 & 0.542 & 0.656 & 0.649 & 0.410 & 0.651 & 0.627 & 0.480 & 0.373 & 0.412 & 0.735 & 0.539 & \textbf{0.541} \\
1/10000   & 0.566 & 0.431 & 0.419 & 0.619 & 0.542 & 0.656 & 0.649 & 0.410 & 0.650 & 0.627 & 0.480 & 0.373 & 0.412 & 0.734 & 0.539 & \textbf{0.540} \\
1/100000  & 0.566 & 0.431 & 0.419 & 0.619 & 0.542 & 0.656 & 0.649 & 0.410 & 0.650 & 0.627 & 0.480 & 0.373 & 0.412 & 0.734 & 0.539 & \textbf{0.540} \\
1/1000000 & 0.566 & 0.431 & 0.419 & 0.619 & 0.542 & 0.656 & 0.649 & 0.410 & 0.650 & 0.627 & 0.480 & 0.372 & 0.412 & 0.734 & 0.539 & \textbf{0.540} \\

\bottomrule

  \end{tabularx}
  \caption{
  \textit{mIoU} values for different NFA thresholds for all categories. Tested thresholds are $NFA \leq 1 / 10^i$,  with $i\in [0, 6]$, or what is the same: $\log(NFA) \leq -i$. Theoretically, using the threshold $1 / 10^i$, we would obtain, on average one false alarm every $10 ^i$ images under the null hypothesis. For example, for the first threshold, we expect to have one false alarm per image, while in the last threshold ($1 / 10^6$), we expect one false alarm every one million images. It is evident that in practice nearly all tested threshold values result in the same \textit{mIoU}s readings. By identifying the most significant region for each tree branch, the PFA tree introduced in Section~3.4.2 of the original article makes this threshold extremely robust.
  }
  \label{tab:nfa_thresholds}
\end{table*}

\section{Analysis of the proposed multiple hypothesis testing procedure}

\begin{figure*}[ht]
  \centering
  \includegraphics[width=1\linewidth]
  {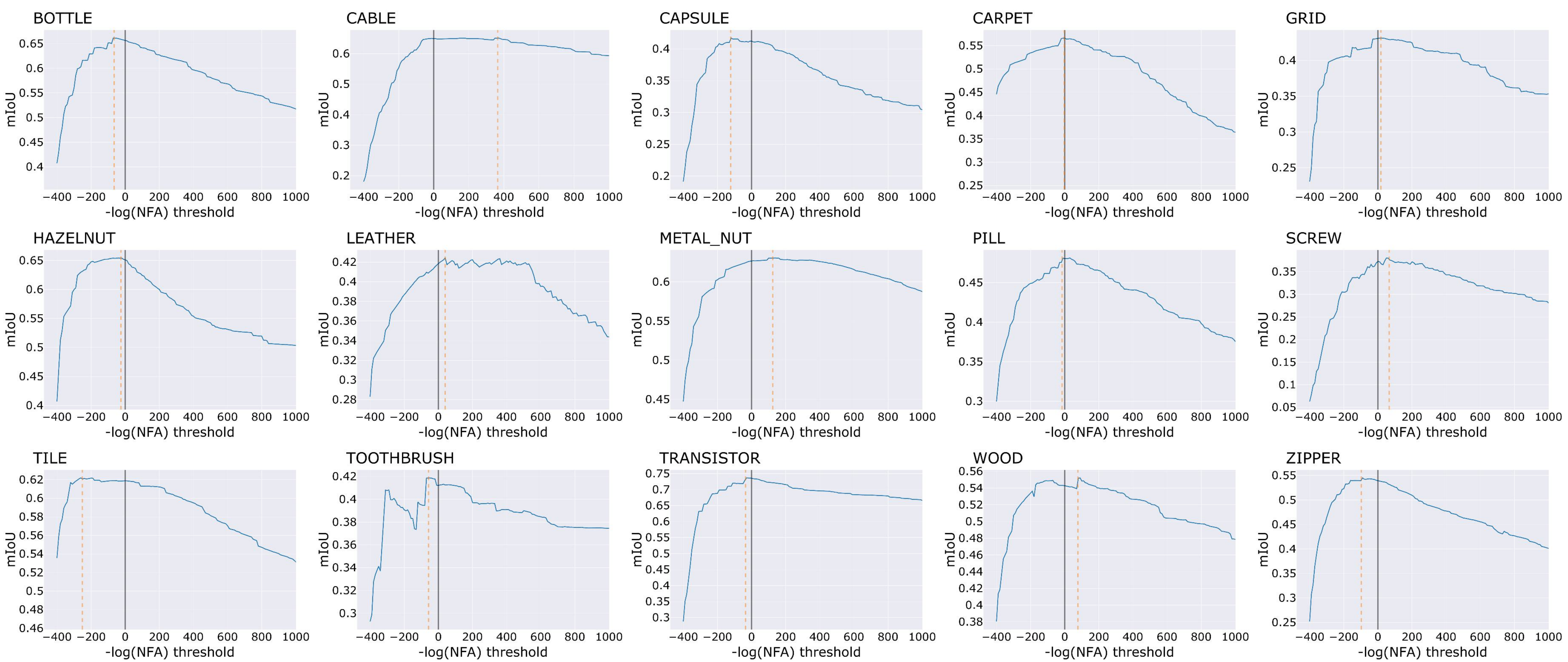}  
   \caption{Segmentation performance (\textit{mIoU}) as a function of the detection threshold over $-\log(\text{NFA})$, for all MVTec categories. The automatic threshold $\log(\text{NFA})=0$ is represented by the black solid line, and the ``oracle'' threshold with the dashed orange line. The latter corresponds to the best possible threshold to maximize \textit{mIoU}. For all categories there is a wide range of thresholds with almost no variation in the performance metric, indicating that the method is robust in the selection of the threshold. Also, the ``oracle'' threshold lays almost always very close to the automatic threshold, supporting the theory behind the \textit{a contrario} methodology.
   Note that for the cases where we obtained the most different thresholds (automatic vs. oracle), the obtained \textit{mIoU} values are actually almost the same.}
   \label{fig:mious_sup}
\end{figure*}

To extend the analysis shown in Figure~7 of the paper, we present Figure~\ref{fig:mious_sup} on this document with a more detailed view. For each category, we show the different performances in terms of \textit{mIoU} obtained with different thresholds over the value of $-\log(\text{NFA})$. The vertical solid line represents the automatic threshold of $\log(\text{NFA})\leq 0$, and the dashed orange line represents the ``oracle'' threshold, i.e., the threshold for which the highest performance is reached.
It can be seen that almost always the automatic threshold lies near the ``oracle'' threshold. Actually, this is not surprising, as the NF is supposed to produce a closed-form representation (a white Gaussian process) of the transformed density, which enables the derivation of automatic thresholds without requiring to ``manually'' learn them. The fact that these thresholds are close is a confirmation that the theoretical densities are close to the sample densities. Moreover, all curves show a wide flat area, indicating that variations in the selection of the threshold do not really affect the segmentation, indicating that this method is robust and in practice parameter-free, as stated in the article. 
Table~\ref{tab:nfa_thresholds} provides a more detailed examination and quantitative assessment of the outcomes across various thresholds. This table exhibits the \textit{mIoU} obtained scores across all MVTec categories, delineated across multiple rows corresponding to different NFA thresholds. Specifically, these thresholds encompass tolerances allowing an average of one false alarm per image, one false alarm per ten images, and progressively extending to one false alarm per one million images. As can be easily seen, particularly when considering the overall average displayed in the last column, the results demonstrate near parity across all thresholds. \\
%

\begin{figure*}[ht]
  \centering
  \includegraphics[width=.72\linewidth]{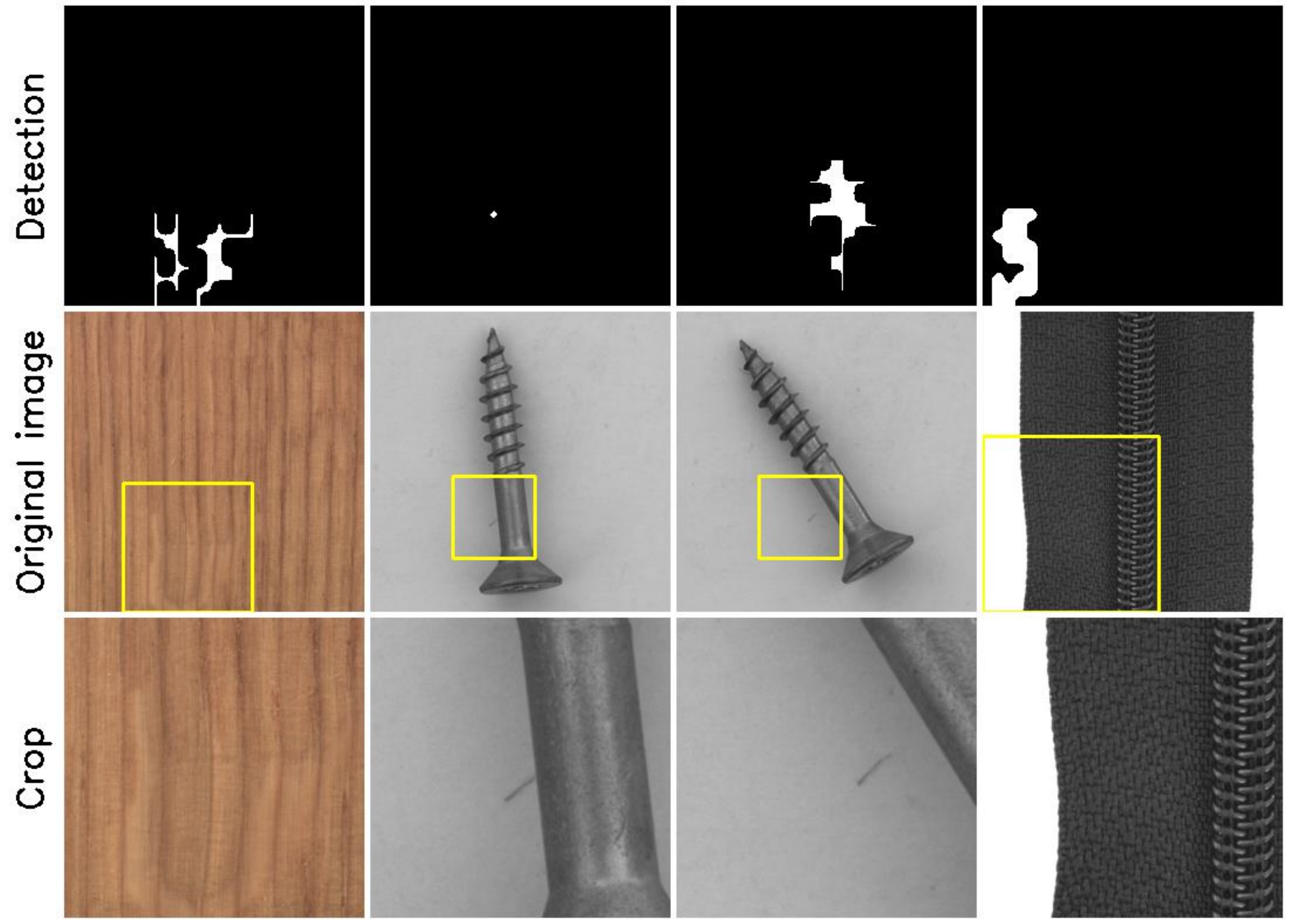}  
   \caption{From all anomaly-free images in the test set of the MVTec dataset, the proposed detection method finds six false alarms, from which four actually correspond to real anomalies that are not labeled as such. This figure shows for these four images the detection using the automatic threshold ($\log(NFA)\leq 0$) in the first row, the original image in the middle row, and a crop containing the anomaly in the last row to ease the visualization.}
   \label{fig:false_errors}
\end{figure*}

Finally, we present illustrative examples to evaluate the empirical false alarm rate. Our evaluation involved conducting detections on all anomaly-free test images, a total number of 467 images. Among these, six instances were flagged as false alarms, but notably, four of these detections corresponded to genuine anomalies that had not been labeled as such. Those images are presented in Figure~\ref{fig:false_errors}. In summary, we obtained only two real false alarms from all 467 test images. It is noteworthy that these two false alarms encompass multiple sets of detections, as previously discussed, indicating a potential underlying convergence of the actual average false alarm rate towards the expected number predicted by theory.

